\documentclass[accepted,x11names]{uai2022} 

\usepackage[american]{babel}

\usepackage[numbers]{natbib} 
\usepackage{mathtools} 
\usepackage{booktabs} 
\usepackage{tikz} 
\usepackage{array}
\usepackage{ragged2e}
\usepackage{xr-hyper} 
\newcolumntype{P}[1]{>{\RaggedRight\hspace{0pt}}p{#1}}

\usepackage{lipsum}
\usepackage{amsmath}
\usepackage{amssymb}
\usepackage{amsthm}
\usepackage{wrapfig}
\usepackage[utf8]{inputenc} 
\usepackage[T1]{fontenc}    
\usepackage{url}            
\usepackage{booktabs}       
\usepackage{amsfonts}       
\usepackage{nicefrac}       
\usepackage{microtype}      
\usepackage{xcolor}         
\usepackage{graphicx}
\usepackage{placeins}
\usepackage{enumitem}
\usepackage{wrapfig}
\usepackage{subcaption}
\usepackage[noend]{algorithm2e}

\newcommand{\E}[0]{\mathbb{E}}
\newcommand{\LL}[0]{\mathcal{L}}
\newcommand{\UU}[0]{\mathcal{U}}
\newcommand{\grad}[0]{\nabla_{\theta}}

\makeatletter
\newcommand{\removelatexerror}{\let\@latex@error\@gobble}
\makeatother

\theoremstyle{plain}
\newtheorem{theorem}{Theorem}[section]

\newtheorem{lemma}[theorem]{Lemma}

\theoremstyle{definition}

\theoremstyle{remark}

\hypersetup{
    colorlinks=true,
    urlcolor=black,
    linkcolor=cyan,
    citecolor=SpringGreen3,
    linkbordercolor=white,
}

\title{Recursive Monte Carlo and Variational Inference with Auxiliary Variables}

\author[1]{\href{mailto:<alexlew@mit.edu>?Subject=Your UAI 2022 paper}{Alexander~K.~Lew}{}}
\author[1]{\href{mailto:<marcoct@mit.edu>?Subject=Your UAI 2022 paper}{Marco~Cusumano-Towner}{}}
\author[1]{\href{mailto:<vkm@mit.edu>?Subject=Your UAI 2022 paper}{Vikash~K.~Mansinghka}{}}
\affil[1]{%
    Massachusetts Institute of Technology\\
    Cambridge, Massachusetts, USA
}
  
\begin{document}
\maketitle

\begin{abstract}
A key design constraint when implementing Monte Carlo 
and variational inference algorithms is that it must 
be possible to cheaply and exactly evaluate the marginal 
densities of proposal distributions and variational families.
This takes many interesting 
proposals off the table, such as those
based on involved simulations or stochastic optimization.
This paper broadens the design space, by presenting 
a framework for applying Monte Carlo and variational 
inference algorithms when proposal densities cannot 
be exactly evaluated. Our framework, \textit{recursive
auxiliary-variable inference} (RAVI), instead approximates 
the necessary densities using \textit{meta-inference}: 
an additional layer of Monte Carlo or variational inference,
that targets the proposal, rather than the model.
%
RAVI generalizes and unifies several existing methods for 
inference with expressive approximating families, 
which we show correspond to specific choices 
of meta-inference algorithm, and provides new theory for analyzing their bias and variance.
We illustrate RAVI's design framework and theorems by using them to analyze and improve upon \citet{salimans2015markov}'s Markov Chain Variational Inference, and to design a novel sampler for Dirichlet process mixtures, achieving state-of-the-art results on a standard benchmark dataset from astronomy and on a challenging data-cleaning task with Medicare hospital data.
  
\end{abstract}

\newcommand\mycommfont[1]{\scriptsize\ttfamily\textcolor{darkgray}{#1}}
\SetCommentSty{mycommfont}

\begin{table*}[t]
    \begin{tabular}{P{0.28\linewidth}P{0.32\linewidth}P{0.35\linewidth}}
    \textbf{Monte Carlo or variational inference algorithm} & \textbf{Distributions that no longer need fast exact density evaluators} & \textbf{Example applications} \\
    Importance Sampling~\citep{glynn1989importance} (Alg.~\hyperref[alg:alg1]{1}, Appendix~\ref{sec:sir-example})  & proposal $q(x; y)$                                                                & Nested IS~\citep{naesseth2015nested} (Appendix~\ref{sec:nsmc-example}), Agglomerative Monte Carlo (Section~\ref{sec:examples}, RAVI strategy~\hyperref[infstrat:agglom]{2}), Annealed IS~\citep{neal2001annealed} (Appendix~\ref{sec:ais-example})   \\
    Particle Filtering~\citep{djuric2003particle} (Appendix~\ref{sec:smc-example}) & initial proposal $q_0(x_0; y_0)$, step proposals $q_t(x_i \mid x_{t-1}, y_{t})$        &    Nested SMC~\citep{naesseth2015nested} (Appendix~\ref{sec:nsmc-example}), SMC$^2$~\citep{chopin2013smc2} (Appendix~\ref{sec:smcsq-example})   \\
    Del-Moral SMC~\citep{del2006sequential} (Appendix~\ref{sec:smc-example})      & initial proposal $q_0(x_0)$, forward kernels $K_t(x_t \mid x_{t-1})$, reverse kernels $L_t(x_{t-1} \mid x_t)$, targets $\tilde\pi_t(x)$                                                                                &                                           \\
    Black-Box Variational Inference~\citep{ranganath2014black} (Alg.~\hyperref[alg:alg3]{3}) & variational family $q_\theta(x; y)$ & IWAE~\citep{burda2015importance} (Appendix~\ref{sec:iwae-example}), MCVI~\citep{salimans2015markov} (Section~\ref{sec:overview}, Appendix~\ref{sec:ham-example}), Variational SMC~\citep{naesseth2018variational} (Appendix~\ref{sec:vsmc-example})\\
    Amortized Variational Inference~\citep{le2017inference} (Alg.~\hyperref[alg:alg4]{4}) & variational family $q_\theta(x; y)$ & Amortized Rejection Sampling~\citep{naderiparizi2019amortized} (Appendix~\ref{sec:amrej-example})\\
    Metropolis-Hastings  (Alg.~\hyperref[alg:alg5]{5})  &   transition proposal $q(x'; x)$                                                                                 &             pseudo-marginal ratio MH~\citep{andrieu2018utility}                              \\
    Hierarchical Variational Inference~\citep{ranganath2016hierarchical} & variational family $q_\theta(z, x; y)$, reverse proposal $r_\theta(z; x, y)$ & Importance-Weighted HVI~\citep{sobolev2019importance}, RAVI-MCVI (Sections~\ref{sec:overview} and \ref{sec:examples}, RAVI strategy~\hyperref[infstrat:mcvi]{1})
    \end{tabular}
    \caption{RAVI generalizes many algorithms for Monte Carlo and variational inference, by allowing 
    practitioners to choose proposals, variational families, and intermediate targets for which 
    exact density evaluators are not available. In the ``example applications'' column, we list 
    both novel examples of algorithms that exploit this degree of freedom (e.g., the Agglomerative Monte Carlo algorithm we develop in Section~\ref{sec:examples}), and algorithms from the literature that --- as we show in Appendix~\ref{sec:appendix-examples} --- can be viewed as instances of simpler algorithms, but with certain sophisticated proposals whose density RAVI estimates.}
\end{table*}

\section{INTRODUCTION}
\label{sec:intro}
Monte Carlo and variational inference algorithms 
are the workhorses of modern probabilistic inference,
a fundamental problem with applications in many disciplines~\citep{murphy2012machine}.
A key challenge in applying these algorithms is the design of \textit{proposal distributions} (in VI, variational families), which can greatly affect their performance~\citep{chatterjee2018sample}. A good proposal should incorporate any knowledge the practitioner might have about the shape of the posterior; however, this goal is often in tension with the requirement that a proposal's marginal density be analytically tractable, in order to compute importance weights, MCMC acceptance probabilities, or gradient updates for VI. The challenge is that proposal distributions that are simple enough to admit exact density evaluators may not be flexible enough to solve real-world posterior inference problems.

In this paper, we present a new framework, called \textit{Recursive Auxiliary-Variable Inference} (RAVI), for incorporating more complex proposals, without exact marginal density evaluators, into standard Monte Carlo and VI algorithms.
The key idea is to approximate the 
proposal densities using \textit{meta-inference}~\citep{cusumano2017aide}: an additional layer of Monte Carlo or variational inference targeting the proposal, rather than the model. RAVI generalizes and unifies several existing methods for inference with expressive proposals~\citep{salimans2015markov,ranganath2016hierarchical,sobolev2019importance}, which we show correspond to specific choices of meta-inference algorithm (see Appendix~\ref{sec:appendix-examples} for 10 examples).

\textbf{Contributions.} Our key contributions are:
\begin{itemize}
\item the RAVI framework, including new recursive algorithms for IS, VI, SMC, and MH using proposals without exact marginal density evaluators (Sections~\ref{sec:overview} \&~\ref{sec:inference-algs});
\item theorems characterizing the impact of RAVI's estimated densities on inference quality (sampler variance, or tightness of variational bounds) (Section~\ref{sec:theory}); and
\item two extended examples of RAVI's application to algorithm design and analysis: (1) a novel variant of~\citet{salimans2015markov}'s Markov Chain Variational Inference (MCVI) algorithm that, unlike vanilla MCVI, scales to handle proposals incorporating long MCMC chains; and (2) a novel sampler for Dirichlet process mixtures that uses a randomized agglomerative clustering algorithm as a proposal, outperforming strong baselines on a standard benchmark from astronomy~\citep{drinkwater2004large} and a challenging Medicare data cleaning problem~\citep{MedicareHosp,lew2021pclean}.
\end{itemize}

\section{RECURSIVE AUXILIARY-VARIABLE INFERENCE}
\label{sec:overview}

In this section, we introduce the RAVI framework in the context of a running example: 
we incorporate a chain of MCMC steps into a proposal, so that it can more accurately approximate a
posterior distribution. Our approach generalizes~\citet{salimans2015markov}'s Markov Chain Variational Inference (MCVI) algorithm, and fixes a flaw that prevents it from scaling to longer MCMC chains.



\textbf{An expressive proposal based on MCMC.} 
Let $p(x, y)$ be a latent-variable model and  
$y$ an observation. Suppose we wish to approximate  
$p(x \mid y)$ using an expressive proposal $q(x)$, that
generates an initial location $x_0$ from a simple 
parametric distribution $q_0$, then 
iterates $M$ steps of an MCMC kernel $T$:\footnote{Why incorporate $M$ MCMC steps into a proposal $q$, rather than simply running MCMC?
Several reasons: (1) if we use $q$ as an importance sampling proposal, the importance weights are unbiased 
estimates of the marginal likelihood $p(y)$, which we can use to evaluate our model; (2) if we use $q$ as a variational family, we can optimize the ELBO to learn parameters of the initial proposal or the MCMC transition kernel; and (3) if we generate many importance sampling particles using $q$, their importance weights can in theory correct for the bias of finite-sample MCMC.}
$$q(x) = \int q_0(x_0) \left(\prod_{i=1}^M T(x_{i-1}\rightarrow x_i)\right) \delta_{x_M}(x) \text{d}x_{0:M}.$$
Even when $q_0$ is a poor approximation to $p(x \mid y)$, $q(x)$ may be close to the posterior, if $M$ is sufficiently high. However, because the density $q(x)$ cannot be efficiently evaluated, we cannot use $q(x)$ as a proposal within importance sampling (we have no way to evaluate the importance weight $\frac{p(x, y)}{q(x)}$), nor as a variational family in VI (we cannot estimate the ELBO $\mathcal{L} = \mathbb{E}_{x \sim q} [\log \frac{p(x,y)}{q(x)}]$ or its gradient, making it impossible to learn $p$'s or $q$'s parameters).%
\begin{figure}[t]
    \centering
    \includegraphics[width=0.9\linewidth]{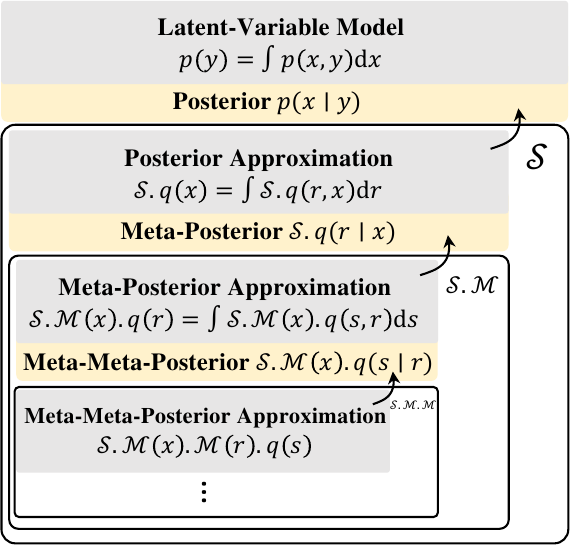}
    \caption{Structure of a RAVI inference strategy $\mathcal{S}$ targeting the posterior $p(x \mid y)$ of a latent-variable model. 
    The proposal $\mathcal{S}.q(x) = \int \mathcal{S}.q(r, x) \text{d}r$ has an intractable marginal density, so the strategy also specifies a \textit{meta-inference} strategy $\mathcal{S}.\mathcal{M}$ that targets $\mathcal{S}.q(r \mid x)$. Nesting continues until the $q$
    approximation at some layer has a tractable density, at which 
    point no further meta-inference is needed.}
    \label{fig:nested}
\end{figure}

\textbf{Approximating proposal densities with meta-inference.} RAVI's goal is to enable inference even when we cannot compute the marginal densities of our proposals and variational families exactly. To apply RAVI, we must specify not just the proposal itself but also a \textit{meta-inference} algorithm, bundled with the proposal into an \textit{inference strategy}:

\textbf{Definition.} An \textit{inference strategy $\mathcal{S}$ targeting $\pi$} specifies:
\begin{itemize}
    \item a posterior approximation $\mathcal{S}.q(x) \approx \pi(x)$\footnote{To simplify the exposition, we assume that if an inference strategy $\mathcal{S}$ targets $\pi$, then the approximation $\mathcal{S}.q$ is \textit{mutually absolutely continuous} with $\pi$, i.e. the measure-zero events under $\pi$ are exactly the same as those under $\mathcal{S}.q$. This requirement can be relaxed somewhat; see Appendix~\ref{sec:even-odd}.} that either has an efficient density evaluator, or is the marginal distribution of a joint distribution with a tractable density, i.e. $\mathcal{S}.q(x) = \int \mathcal{S}.q(r, x) \text{d}r$, and,
    
    \item if $\mathcal{S}.q$'s marginal density cannot be efficiently evaluated, a \textit{meta-inference strategy} $\mathcal{S}.\mathcal{M}$, assigning to each value of $x$ an inference strategy $\mathcal{S}.\mathcal{M}(x)$ targeting $\mathcal{S}.q(r \mid x)$.
\end{itemize}

Figure~\ref{fig:nested} illustrates the recursive 
structure of an inference strategy. 
The key novelty is the inclusion of \textit{meta-inference}, in the form of 
\textit{meta-posterior approximations}: 
additional proposals that the user specifies for 
inferring auxiliary variables introduced by existing proposal distributions. 
In our running example, we take $\mathcal{S}.q(x)$ to be our MCMC-based posterior approximation: it lacks a tractable density, but is the marginal of a tractable joint density $\mathcal{S}.q(x_{0:M}, x)$ over entire MCMC traces. A \textit{meta-posterior approximation}, then, is a probability distribution $\mathcal{S}.\mathcal{M}(x).q(x_{0:M})$ that approximates the \textit{meta-posterior} $\mathcal{S}.q(x_{0:M} \mid x)$: the distribution over traces of the MCMC chain, given 
the final location $x$. 

The meta-posterior approximations enable RAVI to estimate the intractable marginal density of the top-level posterior approximation, to compute weights and gradients:

\textit{In Monte Carlo:} If $\mathcal{S}.q(x) = \int \mathcal{S}.q(r, x) \text{d}r$ is intended for use as a Monte Carlo proposal, RAVI uses meta-inference to obtain an unbiased estimate of $\frac{1}{\mathcal{S}.q(x)}$ (Algorithm~\hyperref[alg:alg2]{2}), which is then multiplied by $p(x, y)$ to estimate the importance weight $\frac{p(x, y)}{\mathcal{S}.q(x)}$. This process relies on the \textit{harmonic mean identity}~\citep{newton1994approximate}, that for any meta-posterior approximation $h$, $$\mathbb{E}_{\mathcal{S}.q(r \mid x)}\left[\frac{h(r)}{\mathcal{S}.q(r, x)}\right] = \frac{1}{\mathcal{S}.q(x)}\mathbb{E}\left[\frac{h(r)}{\mathcal{S}.q(r \mid x)}\right] = \frac{1}{\mathcal{S}.q(x)}.$$
(Harmonic mean estimators are infamous for having potentially infinite variance, but only when $h$ is set to a broad prior; we give a general analysis of the variance of RAVI's importance weights in Section~\ref{sec:theory}.)

\textit{In Variational Inference:} If $\mathcal{S}.q(x) = \int \mathcal{S}.q(r, x) \text{d}r$ is intended as a variational family, then RAVI uses the meta-posterior approximation to formulate an \textit{upper bound} on $\log \mathcal{S}.q(x)$: for any meta-posterior approximation $h(r)$,
$$\log \mathcal{S}.q(x) \leq \mathcal{U}(x) := \mathbb{E}_{\mathcal{S}.q(r \mid x)}[\log \mathcal{S}.q(r, x) - \log h(r)].$$
This follows from Jensen's inequality, and the harmonic 
mean identity from above. 
With this upper bound in hand, we formulate a surrogate ELBO $\mathcal{L}_\mathcal{S} = \mathbb{E}_{\mathcal{S}.q(x)}[\log p(x, y) - \mathcal{U}(x)] \leq \mathcal{L}$, which we can tractably estimate and optimize via stochastic gradient descent (Algorithm~\hyperref[alg:alg3]{3}).

In Section~\ref{sec:inference-algs}, we show how similar estimators can be built up recursively when the meta-posterior approximations themselves have intractable marginal densities.

\textbf{A meta-inference strategy that recovers the MCVI objective~\citep{salimans2015markov}.} In our running example, where the auxiliary randomness $r$ is a trace $x_{0:M}$ of locations visited by MCMC, one option for meta-inference is to learn neurally parameterized 
reverse Markov kernels $R_i(x_{i+1} \rightarrow x_i)$, and apply them 
in sequence to infer a plausible trace of MCMC steps leading to the 
final location $x$:%
$$\mathcal{S}.\mathcal{M}(x).q(x_{0:M}) = \delta_{x}(x_M) \prod_{i=0}^{M-1} R_i(x_{i+1} \rightarrow x_i).$$%
This approximation to $\mathcal{S}.q(x_{0:M} \mid x)$ has a tractable 
density, and so completely specifies the meta-inference strategy $\mathcal{S}.\mathcal{M}$; there is no need to specify a \textit{meta-}meta-inference 
strategy. Given $\mathcal{S}$, RAVI optimizes the surrogate objective
$\mathcal{L}_\mathcal{S} = \mathbb{E}_{x \sim \mathcal{S}.q}[\log p(x, y) - \mathcal{U}_{\mathcal{S}.\mathcal{M}(x)}]$, where
$$\mathcal{U}_{\mathcal{S}.\mathcal{M}(x)} = \mathbb{E}_{x_{0:M} \sim \mathcal{S}.q(x_{0:M} \mid x)}\left[\log \frac{\mathcal{S}.q(x_{0:M}, x)}{\mathcal{S}.\mathcal{M}(x).q(x_{0:M})}\right].$$
For the above choice of $\mathcal{S}.\mathcal{M}$, the RAVI objective $\mathcal{L}_\mathcal{S}$ exactly coincides with the Markov Chain Variational Inference (MCVI) objective of~\citet{salimans2015markov}. In fact, RAVI unifies and generalizes many existing methods; 10 examples are collected in Appendix~\ref{sec:appendix-examples}.

\textbf{Analyzing MCVI within the RAVI framework.} Framing MCVI as a RAVI algorithm lets us analyze it using general theory about RAVI objectives. For example, the relative tightness of the bound $\mathcal{L}_\mathcal{S}$ is controlled by the quality of meta-inference:
$$\mathcal{L} - \mathcal{L}_\mathcal{S} = \mathbb{E}_{\mathcal{S}.q(x)}[KL(\mathcal{S}.q(x_{0:M} \mid x) || \mathcal{S}.\mathcal{M}(x).q(x_{0:M}))].$$
We can use this characterization to analyze the MCVI objective's behavior as $M$ grows, i.e., as MCMC steps are added. Informally, as the MCMC chain begins to mix, the marginal distribution $\mathcal{S}.q(x)$ over the final location of the chain should grow closer to the posterior $p(x \mid y)$, tightening the (intractable) ELBO $\mathcal{L}$. Unfortunately, the meta-inference gap $\mathcal{L} - \mathcal{L}_\mathcal{S}$ \textit{grows} with $M$, unless each kernel $R_i$ exactly captures the local 
posterior $\mathcal{S}.q(x_i \mid x_{i+1})$. (This can be seen as an instance of the well-known \textit{degeneracy problem} of sequential importance sampling~\citep[Proposition 1]{doucet2000sequential}.) As MCMC converges, the rate of improvement in $\mathcal{L}$ slows, and the meta-inference penalty 
for increasing the chain's length eventually outweighs the 
benefit of improving the posterior approximation $\mathcal{S}.q$. The red curves in Figure~\ref{fig:mcvi-experiment} show this phenomenon playing out on two toy targets: we see that $\mathcal{L}_\mathcal{S}$ does 
become tighter as more MCMC steps are added, but only to a point, before the bound begins to \textit{loosen}.

\textbf{Resolving the issue with improved meta-inference.} 
RAVI clarifies that the variational bound loosens with increasing $M$ due to poor meta-inference: as the MCMC chain grows longer, error in the learned backward kernels accumulates.
This analysis also points to 
a solution: use a meta-inference 
strategy $\mathcal{S}.\mathcal{M}$ that \textit{can} scale to 
longer MCMC histories. 

\begin{algorithm}[t]
    \label{infstrat:mcvi}
    \SetAlgoLined\DontPrintSemicolon
    \scriptsize{
    \textbf{RAVI Inference Strategy 1:} RAVI-MCVI\;
    \SetKwFunction{mcvi}{rmcvi($M, K$).q}\SetKwFunction{mcvim}{rmcvi($M, K$).M($x$).q}\SetKwFunction{mcvimm}{rmcvi($M, K$).M($x$).M($x_{0:M})$.q}
    \SetKwProg{infalg}{Posterior Approx.}{}{}
    \SetKwInOut{Infers}{Target of inference}
    \SetKwInOut{Aux}{Auxiliary variables}
    \infalg{\mcvi{}}{
    \Infers{latent variable $x$}
    \Aux{MCMC trace $x_{0:M}$}
    \nl $x_0 \sim q_0$\;
    \nl \For{$i \in 1, \dots, M$}{
        \nl $x_i \sim T(x_{i-1} \rightarrow \cdot)$\;
    }
    \nl \Return{$x_M$}\;}{}
    \setcounter{AlgoLine}{0}
    \SetKwProg{metaalg}{Meta-Posterior Approx.}{}{}
    \metaalg{\mcvim{}}{
    \Infers{MCMC trace $x_{0:M}$}
    \Aux{SMC particles $x_{0:M}^{1:K}$, ancestor indices $a_0, a_{1:M}^{1:K}$}
    \nl \For{$k \in 1,\dots, K$}{
        \nl $(x_M^k, w_M^k, t_k) \gets (x, q_m(x), [x])$\;
    }
    \nl \For{$i \in M-1, \dots, 0$}{
        \nl \For{$k \in 1,\dots,K$}{
            \nl $a_{i+1}^k \sim \text{Discrete}(w_{i+1}^{1:K})$\;
            \nl $x_i^k \sim R_i(x_{i+1}^{a_{i+1}^k} \rightarrow \cdot)$\tcp*{MCVI backward kernel}\;
            \nl $w_i^k \gets \frac{q_i(x_i^k)T(x_i \rightarrow x_{i+1}^{a_{i+1}^k})}{q_{i+1}(x_{i+1}^{a_{i+1}^k})R_i(x_{i+1}^{a_{i+1}^k} \rightarrow x_i^k)}$\;
            \nl $t_k \gets [x_i^k, t_k^{a_{i+1}^k}\dots]$\;
        }
    }
    \nl $a_0 \sim \text{Discrete}(w_{0}^{1:K})$\;
    \nl \Return{$t_{a^0}$}}
    \setcounter{AlgoLine}{0}
    \SetKwProg{metaalg}{Meta-Meta-Posterior Approx.}{}{}
    \metaalg{\mcvimm{}}{
    \Infers{SMC particles $x_{0:M}^{1:K}$, ancestor indices $a_0, a_{1:M}^{1:K}$}
    \Aux{None}
    \nl \For{$i \in 0, \dots, M$}{
        $b_i \sim  \text{Uniform}(1, K)$\;
    }
    \nl \For{$k \in 1,\dots, K$}{
        \nl $(x_M^k, w_M^k) \gets (x, q_m(x)])$\;
    }
    \nl \For{$i \in M-1, \dots, 0$}{
        \nl \For{$k \in 1,\dots,K$}{
            \nl \If{$k = b_i$}{
                \nl $(a_{i+1}^k, x_i^k) \gets (b_{i+1}, x_i)$\;
            }
            \nl \Else{
                \nl $a_{i+1}^k \sim \text{Discrete}(w_{i+1}^{1:K})$\;
                \nl $x_i^k \sim R_i(x_{i+1}^{a_{i+1}^k} \rightarrow \cdot)$\;
            }
            \nl $w_i^k \gets \frac{q_i(x_i^k)T(x_i \rightarrow x_{i+1}^{a_{i+1}^k})}{q_{i+1}(x_{i+1}^{a_{i+1}^k})R_i(x_{i+1}^{a_{i+1}^k} \rightarrow x_i^k)}$\;
        }
    }
    \nl $a_0 \gets b_0$\;
    \nl \Return{$(a_0, a_{1:M}^{1:K}, x_{0:M}^{1:K})$}}
    }
    \vspace{-5mm}
\end{algorithm}

A standard approach 
to resolving the degeneracy problem when inferring 
sequences of latent variables is \textit{sequential 
Monte Carlo} (SMC)~\citep{del2006sequential}. SMC tracks 
$K$ hypotheses about a latent sequence, periodically 
weighting the hypotheses and \textit{resampling}, to 
clone promising particles and cull poor ones. Using 
RAVI, we can use SMC for 
\textit{meta-inference}: we choose $\mathcal{S}.\mathcal{M}(x).q(x_{0:M})$ to generate a collection of $K$ possible backward 
MCMC trajectories, using SMC, before selecting one to return.
This meta-posterior approximation is shown in RAVI Inference Strategy~\hyperref[infstrat:mcvi]{1}. 



This algorithm does not itself have a tractable marginal density: computing $\mathcal{S}.\mathcal{M}(x).q(x_{0:M})$
would require large sums over the resampling variables and intractable integrals over the particle collection. But 
this is where RAVI's recursive structure comes into play:
a meta-inference strategy may use an intractable meta-posterior
approximation, so long as 
we attach a \textit{meta-}meta-inference strategy $\mathcal{S}.\mathcal{M}(x).\mathcal{M}(x_{0:M}).q(a_0, a_{1:M}^{1:K}, x_{0:M}^{1:K})$. In this case meta-meta-inference must infer the auxiliary variables of SMC (ancestor variables and unchosen trajectories), given the final chosen trajectory $x_{0:M}$. For this we can use the \textit{conditional SMC} algorithm~\citep{andrieu2010particle}, 
which runs SMC, with the same auxiliary variables, but constrained to ensure that one of the $K$ particles 
traces the observed trajectory $x_{0:M}$. Because cSMC 
introduces no new auxiliary variables, it has a 
tractable density, and there is no need to specify 
a fourth layer of meta-inference. The full 
tower of posterior approximations is given in RAVI Inference Strategy~\hyperref[infstrat:mcvi]{1}.

In Section~\ref{sec:examples}, we compare MCVI to \texttt{rmcvi}, for different $K$ and $M$. Figure~\ref{fig:mcvi-experiment} shows that meta-inference error is greatly reduced by using SMC, so that the variational bound $\mathcal{L}_\mathcal{S}$ continues to tighten as the MCMC chain grows longer.

\begin{figure*}[t]
\vspace{1mm}
\begin{center}
\textbf{Recursive Monte Carlo Estimation}
\end{center}
\begin{minipage}[t]{0.52\textwidth}
\footnotesize{
\removelatexerror
\vspace{-10pt}
\begin{algorithm}[H]
    \label{alg:alg1}
    \SetAlgoLined\DontPrintSemicolon
    \textbf{Algorithm 1:} RAVI Importance Sampling (\texttt{IMPORTANCE})\;
    \KwIn{unnormalized target $\tilde{\pi}(x) = Z\pi(x)$}
    \KwIn{inference strategy $\mathcal{S}$}
    \KwOut{$(x, \hat{Z})$ properly weighted for $\pi(x)$, s.t. $\mathbb{E}[\hat{Z}] = Z$}
    \nl \uIf{$\mathcal{S}.q$ has a tractable marginal density}{
        \nl $x \sim \mathcal{S}.q$\; 
        \nl $w \gets \frac{1}{\mathcal{S}.q(x)}$\;
    }
    \nl \ElseIf{$\mathcal{S}.q(x) = \int \mathcal{S}.q(r, x)\text{d}r$}{
        \nl $(r, x) \sim \mathcal{S}.q$\;
        \nl $w \gets \texttt{HME}(\mathcal{S}.q(\cdot \mid x), r, \mathcal{S}.\mathcal{M}(x))$\;
    }
    \nl \Return{$(x, w\tilde{\pi}(x))$}\;
\end{algorithm}


}
\end{minipage}\hfill%
\begin{minipage}[t]{0.48\textwidth}
\footnotesize{
\removelatexerror
\vspace{-10pt}
\begin{algorithm}[H]
    \label{alg:alg2}
    \SetAlgoLined\DontPrintSemicolon
    \textbf{Algorithm 2:} RAVI Harmonic Mean Estimation (\texttt{HME})\;
    \KwIn{unnormalized target $\tilde{\pi}(x) = Z\pi(x)$}
    \KwIn{exact sample $x \sim \pi$}
    \KwIn{inference strategy $\mathcal{S}$}
    \KwOut{unbiased estimate $\check{Z}^{-1}$ of $Z^{-1}$}
    \nl \uIf{$\mathcal{S}.q$ has a tractable marginal density}{
        \nl $w \gets \mathcal{S}.q(x)$\;
    }
    \nl \ElseIf{$\mathcal{S}.q(x) = \int \mathcal{S}.q(r, x)\text{d}r$}{
        \nl $(r, w) \gets \texttt{IMPORTANCE}(\mathcal{S}.q(\cdot, x), \mathcal{S}.\mathcal{M}(x))$\;
    }
    \nl \Return{$w/\tilde{\pi}(x)$}\;
\end{algorithm}

}
\end{minipage}
\vspace{-5mm}
\begin{center}
    \textbf{Recursive Variational Objectives and Gradient Estimation}
\end{center}
\vspace{-2mm}
\begin{minipage}[t]{0.52\textwidth}
\footnotesize{
\removelatexerror
\vspace{-10pt}
\begin{algorithm}[H]
    \label{alg:alg3}
    \SetAlgoLined\DontPrintSemicolon
    \textbf{Algorithm 3:} RAVI ELBO and gradient estimator ($\texttt{ELBO}\nabla$)\;
    \KwIn{model $p(x, y)$}
    \KwIn{data $y$}
    \KwIn{inference strategy $\mathcal{S}$}
    \KwOut{unbiased estimates of $\mathcal{L}(p, y, \mathcal{S})$ and of $\nabla_\theta \mathcal{L}(p, y, \mathcal{S})$}
    \nl \If{$\mathcal{S}.q$ has a tractable marginal density}{
        \nl $x \sim \mathcal{S}.q$\;
        \nl$ (\hat{U}, \widehat{\grad}) \gets (\log \mathcal{S}.q(x), \grad \log \mathcal{S}.q(x) \cdot (1 + \log \mathcal{S}.q(x)))$\; 
        \nl $\mathbf{g} \gets \grad \log \mathcal{S}.q(x)$\;
    }
    \nl \ElseIf{$\mathcal{S}.q(x) = \int \mathcal{S}.q(r, x)\text{d}r$}{
        \nl $(r, x) \sim \mathcal{S}.q$\;
        \nl $(\hat{U}, \widehat{\nabla_\theta}, \mathbf{g}) \gets \texttt{EUBO}\nabla(\mathcal{S}.q, x, r, \mathcal{S}.\mathcal{M}(x))$\;
    }
    \nl $\hat{L} \gets \log p(x, y) - \hat{U}$\;
    \nl $\widehat{\grad}'\gets \grad \log p(x, y)
    + \mathbf{g} 
    \log p(x, y)
    - \widehat{\grad}.$\;
    \nl \Return{$(\hat{L}, \widehat{\grad}')$}\;
\end{algorithm}

    
    
    
}
\end{minipage}\hfill%
\begin{minipage}[t]{0.48\textwidth}
\footnotesize{
    \removelatexerror
    \vspace{-10pt}
    \begin{algorithm}[H]
        \label{alg:alg4}
        \SetAlgoLined\DontPrintSemicolon
        \textbf{Algorithm 4:} RAVI EUBO and gradient estimator ($\texttt{EUBO}\nabla$)\;
        \KwIn{model $p(x, y)$}
        \KwIn{data $y$}
        \KwIn{exact sample $x \sim p(x \mid y)$}
        \KwIn{inference strategy $\mathcal{S}$}
        \KwOut{unbiased estimates of $\mathcal{U}(p, y, \mathcal{S})$ and $\nabla_\theta \mathcal{U}(p, y, \mathcal{S})$}
        \KwOut{quantity $\mathbf{g}$ (see Thm. 2)}
        \nl \uIf{$\mathcal{S}.q$ has a tractable marginal density}{
            \nl$ (\hat{L}, \widehat{\grad}) \gets (\log \mathcal{S}.q(x), \grad \log \mathcal{S}.q(x))$\;
        }
        \nl \ElseIf{$\mathcal{S}.q(x) = \int \mathcal{S}.q(r, x)\text{d}r$}{
            \nl $(\hat{L}, \widehat{\nabla_\theta}) \gets \texttt{ELBO}\nabla(\mathcal{S}.q, x, \mathcal{S}.\mathcal{M}(x))$\;
        }
        \nl $\hat{U} \gets \log p(x, y) - \hat{L}$\;
        \nl $\mathbf{g} \gets \nabla_\theta \log p(x, y)$\;
        \nl $\widehat{\grad}'\gets \grad \log p(x, y)
        + \mathbf{g} \cdot \hat{U}
        - \widehat{\grad}$\;
        \nl \Return{$(\hat{U}, \widehat{\grad}', \mathbf{g})$}\;
    \end{algorithm}

    

    

        
}
\end{minipage}
\vspace{-5mm}
\end{figure*}

\textbf{Using the inference strategy within a Monte Carlo algorithm, to estimate marginal likelihoods from MCMC results.} Our inference strategy $\mathcal{S}$ can also be used as proposal within Monte Carlo algorithms, such as importance sampling. In the context of our example, where $\mathcal{S}.q$ incorporates $M$ steps of a Markov chain,
this allows us to assign an \textit{importance weight} to 
each run of the Markov chain.
The weight is an unbiased estimate of the marginal likelihood $p(y)$ of the model;
thus, we can view the algorithm as a way to derive marginal likelihood estimates from MCMC runs, a task of long-standing interest in the Monte Carlo community~\citep{neal2001annealed}. In Section~\ref{sec:examples}, we show that in some settings MCVI compares favorably a standard algorithm for the task, annealed importance sampling (AIS)~\citep{neal2001annealed}.

\section{ALGORITHMS}
\label{sec:inference-algs}

\begin{algorithm}[t]
    \SetAlgoLined\DontPrintSemicolon
    \label{alg:alg5}
    $$\textbf{MCMC}$$
    \footnotesize{
    \hspace{-3mm}\textbf{Algorithm 5:} RAVI Metropolis-Hastings\;
    \KwIn{model $\tilde{\pi}(x) = Z \int \pi(r, x) \text{d}r$}
    \KwIn{proposal $q(x'; x) = \int q(s, x'; x) \text{d}s$}
    \KwIn{family $\mathcal{S}(x)$ of inference strategies targeting $\pi(r \mid x)$}
    \KwIn{family $\mathcal{M}(x, x')$ of inference strategies targeting $q(s \mid x'; x)$}
    \KwIn{initial position $x$ and estimate $\hat{Z}_{x}$ of $\tilde{\pi}(x)$}
    \KwOut{next position $x'$ and estimate $\hat{Z}_{x'}$ of $\tilde{\pi}(x')$}
    \nl $(s, x') \sim q(s, x'; x)$\;
    \nl $w_{x'} \gets {\texttt{HME}}(q(\cdot, x'; x), s, \mathcal{M}(x, x'))$\;
    \nl $(\_, w_x) \gets \texttt{IMPORTANCE}(q(\cdot \mid x; x'), \mathcal{M}(x', x))$\;
    \nl $(\_, \hat{Z}_{x'}) \gets \texttt{IMPORTANCE}(\pi(\cdot \mid x'), \mathcal{S}(x'))$\;
    \nl $u \sim \text{Uniform}(0, 1)$\;
    \nl \If{$u < \text{min}(1, \frac{\hat{Z}_{x'}}{\hat{Z}_x}w_{x'}w_x)$}{
        \nl \Return{$(x', \hat{Z}_{x'})$}\;
    }
    \nl \Else{
        \nl \Return{$(x, \hat{Z}_x)$}\;
    }
    }
\vspace{-3mm}
\end{algorithm}

In this section, we present algorithms for using RAVI inference strategies within Monte Carlo and variational inference algorithms, as proposals and variational families.


\textbf{RAVI for Importance Sampling and SMC.} In importance sampling and SMC algorithms, proposals $q$ are used to (1) generate proposed values $x \sim q$, and (2) compute importance weights $\frac{p(x)}{q(x)}$. But in both IS and SMC, it suffices to produce \textit{unbiased estimates} of $\frac{p(x)}{q(x)}$~\citep{chopin2020introduction}. 
RAVI exploits this degree of freedom to generate proper importance weights even when $q(x)$ is intractable.
Suppose $\tilde{\pi} = Z\pi$ is an unnormalized target density, and $\mathcal{S}$ is a RAVI inference strategy targeting $\pi$. Algorithm~\hyperref[alg:alg1]{1} simulates $x \sim \mathcal{S}.q$ and computes an unbiased estimate $\hat{Z}$ of $\frac{\tilde{\pi}(x)}{\mathcal{S}.q(x)}$:

\textbf{Theorem 1.} \textit{
    Let $\tilde{\pi}(x) = Z\pi(x)$ be an unnormalized target density, and $\mathcal{S}$ an inference strategy targeting $\pi(x)$. Then:
    \begin{itemize}
        \item $\texttt{IMPORTANCE}(\mathcal{S}, \tilde{\pi})$ generates $(x, \hat{Z})$ with $x \sim \mathcal{S}.q$ and $\mathbb{E}[\hat{Z} \mid x] = Z\frac{\pi(x)}{\mathcal{S}.q(x)}$. Furthermore, the unconditional expectation $\mathbb{E}[\hat{Z}(\tilde{\pi}, \mathcal{S})] = Z$.
        \item When $x \sim \pi$, \texttt{HME}($\mathcal{S}, x, \tilde{\pi}$) generates $\check{Z}$ with $\mathbb{E}[{\check{Z}}^{-1}] = Z^{-1}.$
    \end{itemize}
}

When $\mathcal{S}.q$ has a tractable marginal density, Algorithm~\hyperref[alg:alg1]{1} computes an exact importance weight. Otherwise, it calls Algorithm~\hyperref[alg:alg2]{2}, which uses the meta-inference strategy $\mathcal{S}.\mathcal{M}(x)$ to estimate $\frac{1}{\mathcal{S}.q(x)}$. The proof of Theorem 1 is by induction on the level of nesting in the strategy (see Appendix~\ref{sec:proofs}).

\textbf{RAVI for MCMC.} When models or proposals (or both) in a Metropolis-Hastings sampler do not have tractable closed-form densities, RAVI inference strategies enable computation of MH acceptance probabilities (Algorithm~\hyperref[alg:alg5]{5}). Intuitively, to compute the usual Metropolis-Hastings acceptance probability $\alpha = \frac{\tilde{\pi}(x')q(x; x')}{\tilde{\pi}(x)q(x'; x)}$, Algorithm~\hyperref[alg:alg5]{5} estimates the necessary proposal densities, using $\texttt{HME}$ for the forward proposal density that appears in the denominator, and $\texttt{IMPORTANCE}$ for the backward proposal density that appears in the numerator. If necessary, it also uses $\texttt{IMPORTANCE}$ to estimate the new model density $\tilde{\pi}(x')$.
We show the algorithm implements a stationary kernel for $\pi$ in Appendix~\ref{sec:ravi-mcmc}.

\textbf{RAVI for Variational Inference.} Let $p_\theta(x, y)$ be a latent-variable generative model with parameters $\theta$, and $\mathcal{S}_\theta(y)$ is a family of strategies targeting $p_\theta(x \mid y)$. Given a dataset $y$, variational inference can be applied to maximize (a lower bound on) $\log  p_\theta(y)$, and also to optimize parameters of the posterior approximations in $\mathcal{S}_\theta$, to bring them closer (in KL divergence) to their targets.
Let
\begin{align*}
\LL(p, y, \mathcal{S}) &:= \E [ \log \hat{Z}(p(\cdot, y), \mathcal{S}) ] \le \log p(y)\\ \text{and } \UU(p, y, \mathcal{S}) &:= \E [ \log \check{Z}(p(\cdot, y), \mathcal{S}) ] \ge \log p(y),
\end{align*}
where $\hat{Z}(\tilde{\pi}, \mathcal{S})$ is the estimate returned by \texttt{IMPORTANCE} (Alg.~\hyperref[alg:alg1]{1}) on $\mathcal{S}$ and unnormalized target $\tilde{\pi}$,
and $\check{Z}(\tilde{\pi}, \mathcal{S})$ is the inverse of the weight returned from \texttt{HME} (Alg.~\hyperref[alg:alg2]{2}) when run with unnormalized target $\tilde{\pi}$, inference strategy $\mathcal{S}$, and an exact sample $x \sim \pi$. Because $\hat{Z}$ is an unbiased estimate of $p_\theta(y)$, and ${\check{Z}}^{-1}$ is an unbiased estimate of ${p_\theta(y)}^{-1}$, we have by Jensen's inequality that $\LL(p, y, \mathcal{S})$ and $\UU(p, y, \mathcal{S})$ are lower and upper bounds (respectively) on $\log p_\theta(y)$. As such, we can fit the model parameters $\theta$ to data $y$ by minimizing $\UU(p, y, \mathcal{S})$ or maximizing $\LL(p, y, \mathcal{S})$. 

{\it Recursive stochastic gradient estimation.}
$\texttt{ELBO}\nabla$ (Alg.~\hyperref[alg:alg3]{3}) is a procedure for estimating $\LL(p, y, \mathcal{S})$ and its gradient $\grad \LL(p, y, \mathcal{S})$ with respect to the parameters $\theta$ of the model and the strategy. When $(x, y) \sim p(x, y)$, $\texttt{EUBO}\nabla$ (Alg.~\hyperref[alg:alg4]{4}) estimates $\UU(p, y, \mathcal{S})$ and the gradient $\grad\E_{y \sim p}[\UU(p, y, \mathcal{S})]$.
These procedures employ score function estimation of gradients, but 
it is straightforward to incorporate baselines within each procedure to reduce variance.
Depending on $\mathcal{S}$, the reparametrization trick may also be applicable (Appendix~\ref{sec:reparam}). 

\noindent\textbf{Theorem 2.} \textit{
Given a model $p_\theta(x, y)$ and an inference strategy $\mathcal{S}_\theta$ targeting $p_\theta(x \mid y)$,
Alg. 3 yields unbiased estimates of $\LL(p, y, \mathcal{S})$ and of $\grad \LL(p, y, \mathcal{S})$.
Furthermore, when $(x, y) \sim p_\theta$, Alg. 4 yields
(i) $\hat{U}$ such that
$\E [ \hat{U} \mid y ] = \UU(p, y, \mathcal{S})$,
(ii) $\widehat{\grad}$ such that
$\E [ \widehat{\grad} ] = \grad \E_{y \sim p}[\UU(p, y, \mathcal{S})]$,
and (iii) a value $\mathbf{g}$ such that for any function $R$ that does not depend on $\theta$,
$\E [ \mathbf{g} \cdot R(y) ] = \grad \E_{y \sim p_\theta} [ R(y) ]$  if
$\grad \E_{y \sim p_\theta} [ R(y) ]$ is defined.}

In Section 4, we show the tightness of the variational bounds $\LL$ and $\UU$ is given by sums of KL divergences between posterior approximations in $\mathcal{S}_\theta$ and their targets. Thus, optimizing these bounds improves the posterior approximations, either encouraging mass-capturing or mode-seeking behavior.

\section{THEORETICAL ANALYSIS}
\label{sec:theory}

We now present theorems characterizing the quality of RAVI inference: Thm. 3 concerns the variance of weights in a Monte Carlo sampler, and Thm. 4 the tightness of variational bounds. In both cases, error is related to each approximation in the RAVI strategy's divergence to its target posterior.


\textbf{Sampler variance in Monte Carlo.}
Let $\tilde{\pi} = Z\pi$ be an unnormalized target density, and $\mathcal{S}$ an inference strategy targeting $\pi$. As in Section~\ref{sec:inference-algs}, we write $\hat{Z}(\tilde{\pi}, \mathcal{S})$ for the weight returned by \texttt{IMPORTANCE}, and $\text{Var}_{\hat{Z}}(\pi, \mathcal{S})$ for the \textit{relative variance} of the estimator, $\text{Var}(\frac{\hat{Z}(\tilde{\pi}, \mathcal{S})}{Z})$, which does not depend on $Z$ (and therefore is a function of $\pi$, not $\tilde{\pi}$). Similarly, we write $\check{Z}(\tilde{\pi}, \mathcal{S})$ for the reciprocal of the weight returned by \texttt{HME}, run with an input $x \sim \pi$. $\text{Var}_{\check{Z}}(\pi, \mathcal{S})$ is its relative variance, $\text{Var}(\frac{Z}{\check{Z}(\tilde{\pi}, \mathcal{S})})$. 

\noindent\textbf{Theorem 3.} {\it Consider an unnormalized target distribution $\tilde{\pi}(x) = Z\pi(x)$ and an inference strategy $\mathcal{S}$ targeting $\pi(x)$. Then the relative variances of the estimators $\hat{Z}(\tilde{\pi}, \mathcal{S})$ and $\check{Z}(\tilde{\pi}, \mathcal{S})$ are given by the following recursive equations:}
\begin{align*}
\text{Var}_{\hat{Z}}&(\pi, \mathcal{S}) = \chi^2(\pi || \mathcal{S}.q) \, +\\
 & \mathbb{E}_{x \sim \mathcal{S}.q}\left[\left(\frac{\pi(x)^2}{\mathcal{S}.q(x)^2}\right) \cdot \text{Var}_{\check{Z}}(\mathcal{S}.q(\cdot \mid x), \mathcal{S}.\mathcal{M}(x))\right]\\
\text{Var}_{\check{Z}}&(\pi, \mathcal{S}) = \chi^2(\mathcal{S}.q || \pi) + \\
& \mathbb{E}_{x \sim \pi}\left[\left(\frac{\mathcal{S}.q(x)^2}{\pi(x)^2}\right) \cdot \text{Var}_{\hat{Z}}(\mathcal{S}.q(\cdot \mid x), \mathcal{S}.\mathcal{M}(x))\right]
%
\end{align*}
{\it When $\mathcal{S}.q$ is tractable, the second term of each sum is 0.}


\textbf{Tightness of variational bounds.}
In VI, the tightness of the variational bounds $\mathcal{L}$ and $\mathcal{U}$ can be characterized as a sum of a KL divergence and a term measuring meta-inference error. 
The random variables $\hat{\mathcal{L}}$ and $\hat{\mathcal{U}}$ returned by $\texttt{ELBO}\grad$ and $\texttt{EUBO}\grad$, respectively, are unbiased estimators of $\mathcal{L}(p, y, \mathcal{S})$ and $\mathcal{U}(p, y, \mathcal{S})$, and so can also be viewed as
 \textit{biased} estimators of $\log p(y)$. Writing their bias as $\text{Bias}_\mathcal{L}(p, y, \mathcal{S})$ (and similarly for $\mathcal{U}$), we have:

\noindent\textbf{Theorem 4.}
{\it Consider a joint distribution $p(x, y)$ and an inference strategy $\mathcal{S}$ targeting $p(x \mid y)$. Then the following equations give the bias of ${\hat{\LL}}$ and ${\hat{\UU}}$ as estimators of $\log p(y)$:}
\begin{align*}
\text{Bias}_\mathcal{L}(p, y, \mathcal{S}) =&\, -\text{KL}(\mathcal{S}.q || p(\cdot \mid y)) \\ 
&-\mathbb{E}_{x \sim \mathcal{S}.q}[\text{Bias}_\mathcal{U}(\mathcal{S}.q, x, \mathcal{S}.\mathcal{M}(x))]\\
\text{Bias}_\mathcal{U}(p, y, \mathcal{S}) =&\, \text{KL}(p(\cdot \mid y) || \mathcal{S}.q)\\ 
&\,\,\,-\mathbb{E}_{x \sim p(\cdot \mid y)}[\text{Bias}_\mathcal{L}(\mathcal{S}.q, x, \mathcal{S}.\mathcal{M}(x))]
\end{align*}
{\it where the second term in each equation is 0 when $\mathcal{S}.q$ has a tractable marginal density.}

Maximizing $\LL$, or minimizing $\UU$, also minimizes these KL divergences. In particular, maximizing $\LL(p, y, \mathcal{S})$ minimizes a `mode-seeking' KL from $\mathcal{S}.q$ to the posterior, whereas minimizing $\mathbb{E}_{y \sim p}[\UU(p, y, \mathcal{S})]$, e.g. by following the gradients of Alg.~\hyperref[alg:alg4]{4}, implements amortized variational inference, and encourages $\mathcal{S}.q$ to cover the mass of the posterior.

\textbf{Inference and Meta-Inference.} In both Theorems 3 and 4, the first term of the sum is a divergence between $\mathcal{S}.q(x)$, the intractable posterior approximation, and the actual target posterior $p(x \mid y)$. The other term measures the \textit{expected} quality of meta-inference. Thus the overall error of a RAVI algorithm can be understood as decomposing cleanly into (1) the mismatch between the posterior and the intractable proposal, and (2) the error introduced by meta-inference. 

\section{EXPERIMENTS}
\label{sec:examples}

\begin{figure}
    \begin{minipage}[t]{0.48\linewidth}
        \scriptsize{$\mathcal{S}_{\texttt{agglom}}.q$: \textbf{Randomized Clustering}}
        \includegraphics[width=\linewidth]{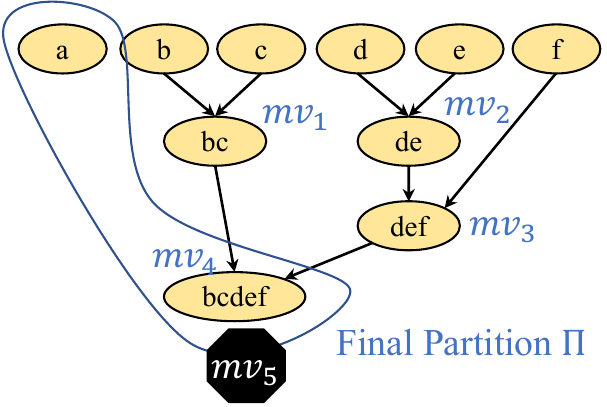}
    \end{minipage}%
    \hfill%
    \begin{minipage}[t]{0.48\linewidth}
        \scriptsize{$\mathcal{S}_{\texttt{rmcvi}}.q$: \textbf{Langevin Monte Carlo}}
        \includegraphics[width=\linewidth]{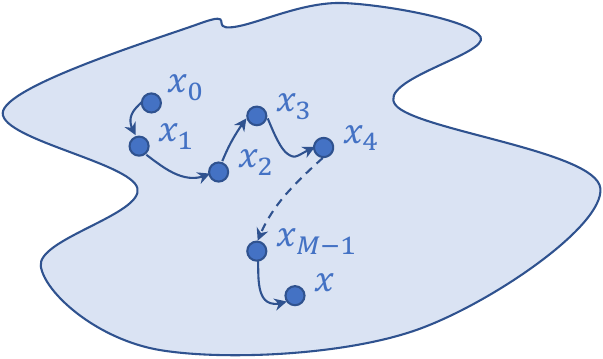}
    \end{minipage}
    \scriptsize{$$\mathcal{S}_\texttt{agglom}.q(\Pi) = \sum_{mv_1} \dots \sum_{mv_{N-|\Pi|}} \prod_{i=1}^{N-|\Pi|}q(mv_i \mid mv_{<i})$$}
    \vspace{-3mm}
    \scriptsize{$$\mathcal{S}_\texttt{rmcvi}.q(x) = \int_{\mathbb{R}^M} q_0(x_0) \left(\prod_{i=1}^{M-1} q(x_i \mid x_{i-1})\right) q(x \mid x_{M-1}) \text{d} x_{0:M-1}$$} 

    \caption{Illustrations of the proposals $\mathcal{S}.q$ used in each experiment.
    In each case, $\mathcal{S}.q$ makes a sequence of auxiliary choices before returning a final proposal (the clustering $\Pi$, or the location $x$). Sequential Monte Carlo meta-inference is used to marginalize the sequence of auxiliary variables introduced 
    by the inference process (the merges $\textit{mv}_i$ in $\texttt{agglom}$, and the locations $x_{i}$ in \texttt{rmcvi}).}
\end{figure}

\begin{table}[t]
    \footnotesize{
    \begin{tabular}{p{0.05\textwidth}p{0.07\textwidth}p{0.1\textwidth}p{0.18\textwidth}}
                              & {\bf Inference}              &{\bf Meta-inference}                             &{\bf Meta-meta-inference}  \\
    {\bf\texttt{agglom}} & \textit{Discrete}: $3.0 \times 10^{1928}$ & \textit{Discrete}: $\prod_{n=|\Pi|}^{1000} \frac{n(n-1)}{2}$ & \textit{Discrete}: $(K - 1) \cdot (\prod_{n=|\Pi|}^{1000} \frac{n(n-1)}{2}) \cdot ({1000-|\Pi|})(K-1)!$ \\
    {\bf\texttt{rmcvi}}  & \textit{Continuous}: 1 & \textit{Continuous}: $M$ &\textit{Continuous}: $(K - 1) \cdot M$, \textit{Discrete}: $M(K-1)!$
    \end{tabular}
    }
    \caption{Dimensionality of the continuous latent space, and cardinality of the discrete latent space, 
    over which each layer's inference problem is defined. $K$ is the number of SMC particles used for meta-inference
    (maximum 50 for $\texttt{rmcvi}$, 5 for $\texttt{agglom}$).
    In $\texttt{rmcvi}$, $M$ is the number of MCMC steps (maximum 100 in our experiments).}
    \label{tab:dimension}
\end{table}

\subsection{Improving MCVI}
\begin{figure*}[t]
    \includegraphics[width=0.33\linewidth]{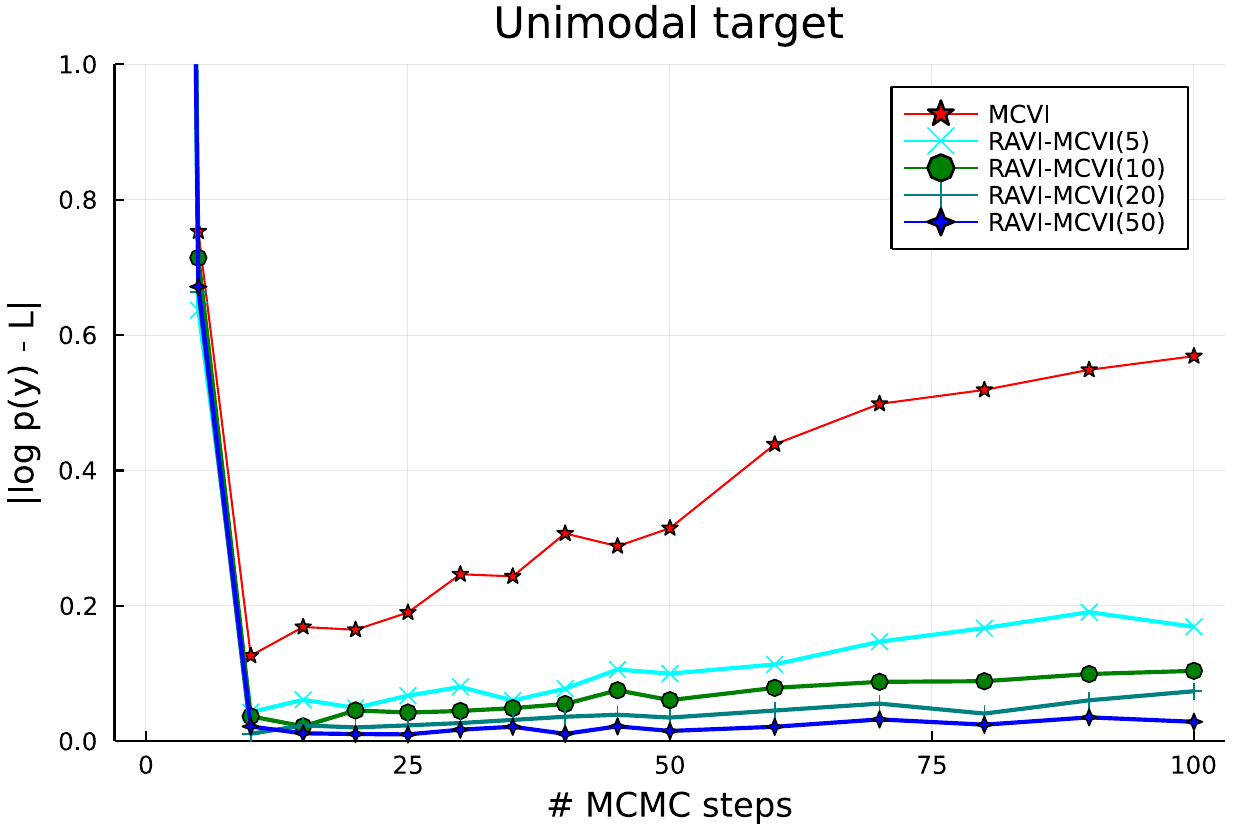}
    \includegraphics[width=0.33\linewidth]{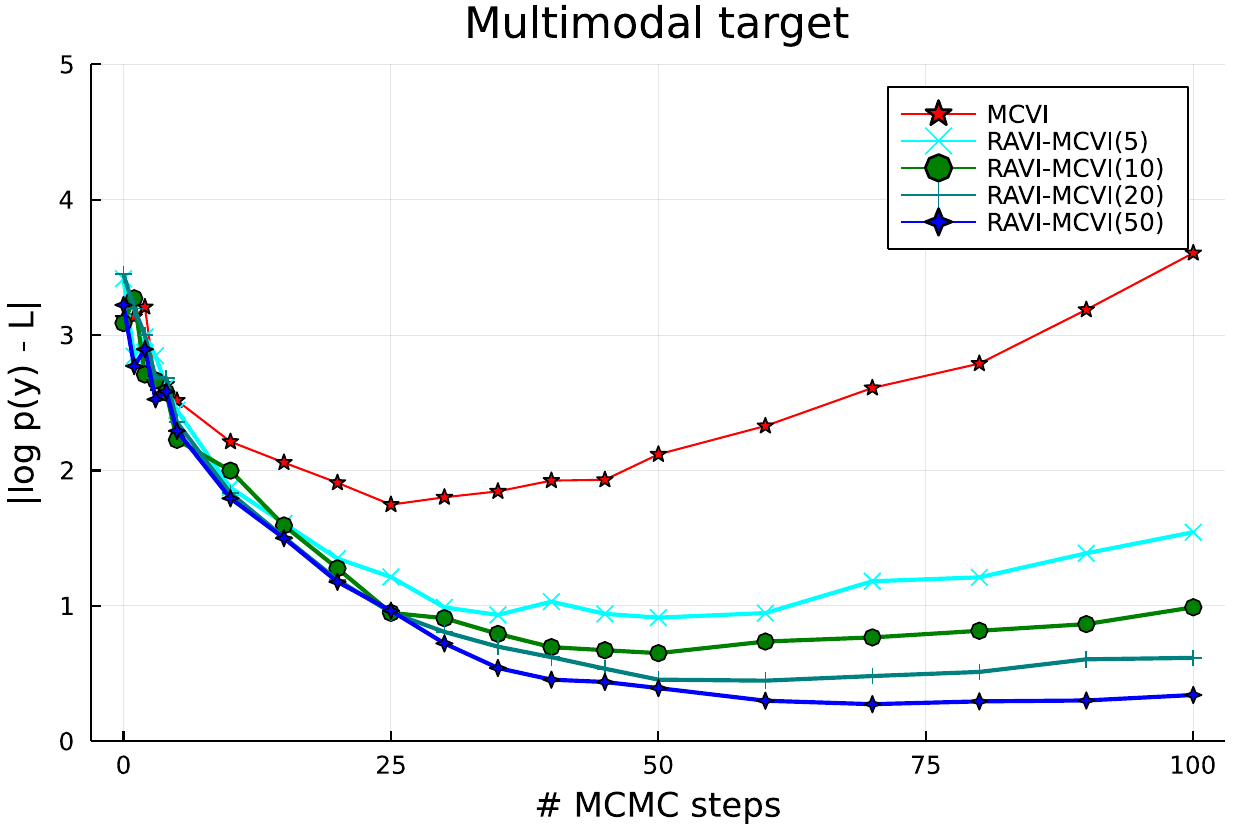}
    \includegraphics[width=0.33\linewidth]{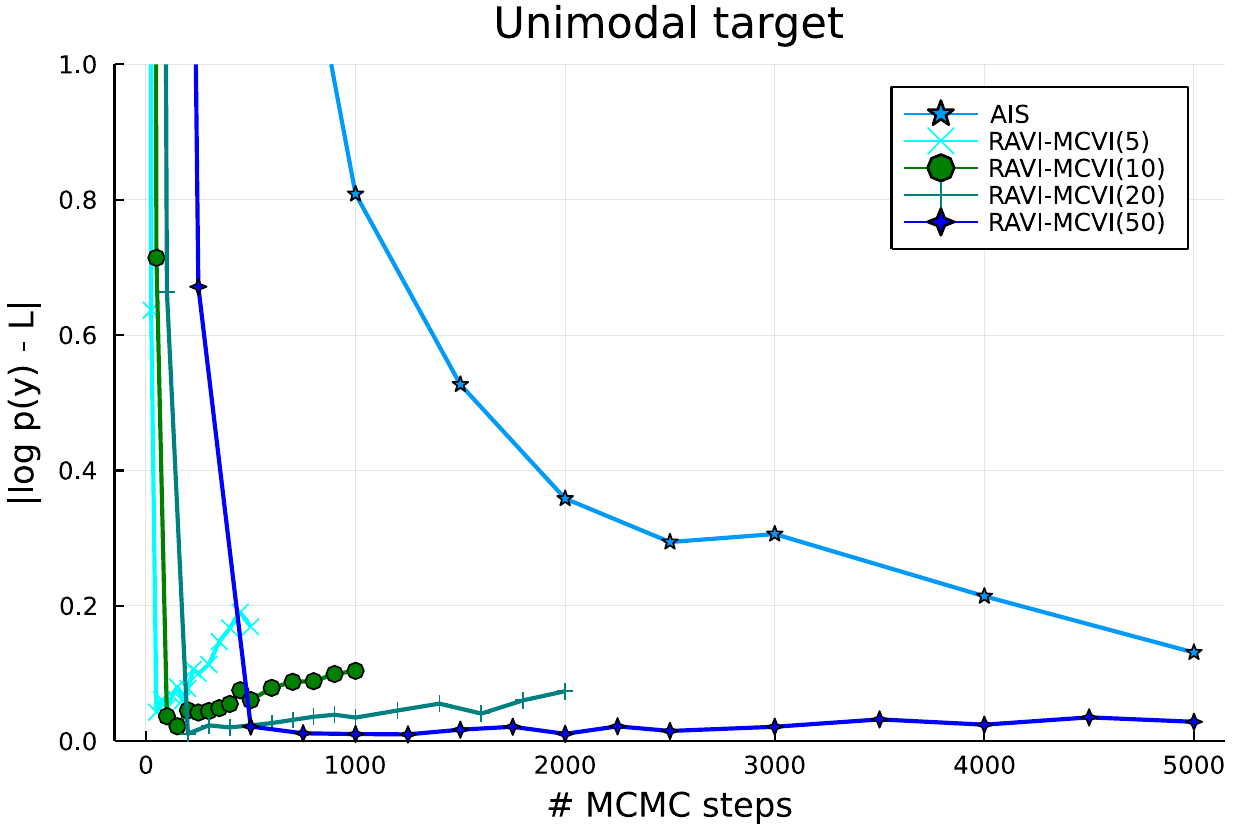}
    \caption{Improving Markov Chain Variational Inference with RAVI. \textbf{Left and Middle:} On unimodal and multimodal targets, MCVI begins to degrade after 15-25 steps of MCMC. RAVI-MCVI with sufficiently many particles continues to improve as more MCMC steps are added. 
    \textbf{Right:} When MCMC converges quickly to a reasonable approximation of the posterior, RAVI-MCVI can give more accurate estimates of marginal likelihoods than standard techniques such as AIS. The $x$ axis of this plot counts total MCMC steps simulated, whether as part of inference or meta-inference; for RAVI-MCVI($K$), this is $KM$, where $M$ is the length of the forward Markov chain and $K$ is the number of SMC particles used for meta-inference.}
    \label{fig:mcvi-experiment}
    \vspace{-5mm}
\end{figure*}

In Section~\ref{sec:overview}, we developed a 
variant of~\citet{salimans2015markov}'s MCVI algorithm 
that used SMC for meta-inference.
In Figure~\ref{fig:mcvi-experiment},
we compare vanilla MCVI to the RAVI variant,
with varying $K$ (number of particles used for 
meta-inference) and $M$ (number of MCMC steps 
in the variational family).

\textbf{Experimental details.}\footnote{Code is available: https://github.com/probcomp/ravi-uai-2022} For the MCMC kernel $T$, we use Langevin 
ascent with step size $0.015$. For the 
meta-inference proposals $R_i(x_{i+1} \rightarrow x_i)$,
we use $\mathcal{N}(x_{i}; f_\mu(x_{i+1}, i), e^{f_{\log \sigma}(x_{i+1}, i)})$, where $f$ is a 4-layer MLP, the step number $i$ 
is encoded as a one-hot vector, and $f$ outputs the mean $\mu$ and log standard deviation $\log\sigma$ for a conditionally Gaussian proposal.
The same $f$ is used for each experiment, and is trained on 
forward rollouts of MCMC (equivalent to using Alg. 3 on $\texttt{rmcvi}$ with $K=1$). The unimodal model is Gaussian with 
$\sigma=0.2$, and the multimodal model is a mixture 
of 3 Gaussians with standard deviations $0.2, 0.3,$ and $2.0$. 
The distributions $q_i$ used for importance weighting during sequential Monte Carlo meta-inference are Gaussians with learned $\mu$ and $\sigma$.

\textbf{Results.} Figure~\ref{fig:mcvi-experiment} plots the gap $\log p(y) - \mathcal{L}$ for each algorithm's variational bound $\mathcal{L}$. By Theorem 4 this gap is the sum of two terms: $KL(\mathcal{S}.q || p(x \mid y))$ and the expected meta-inference divergence $\mathbb{E}_{x \sim \mathcal{S}.q}[KL(\mathcal{S}.q(x_{0:M} \mid x) || \mathcal{S}.\mathcal{M}(x).q(x_{0:M}))]$. The first term is constant across the algorithms, since they all use the same MCMC-based posterior approximation, so the plots primarily illustrate differences in the quality of meta-inference. MCVI's meta-inference steadily worsens as the chain's length grows, and after 15-25 steps, the meta-inference cost of adding new steps outweighs the benefits to $\mathcal{S}_\texttt{mcvi}.q$, causing the bound $\mathcal{L}$ to loosen. Our variant, with SMC meta-inference, does not suffer the same penalty, and continues to improve as more steps are added. As discussed in Section~\ref{sec:overview}, the same inference strategy (\hyperref[infstrat:mcvi]{\texttt{rmcvi}}) can be used within an importance sampler to derive unbiased marginal likelihood estimates from MCMC runs. The right-hand plot in Figure~\ref{fig:mcvi-experiment} shows that this technique can yield accurate estimates with less computation than AIS~\citep{neal2001annealed}, at least on simple targets. (To fairly account for the computational cost of meta-inference, in the RAVI algorithm we multiply $M$ by $K$ when plotting the total number of MCMC steps.) 
Because the variance of AIS is bounded below by sums of divergences between subsequent pairs of intermediate target distributions, the MCMC chain must be long enough to support a very fine annealing schedule, without large jumps. By contrast, RAVI-MCVI requires 
only that the marginal distribution of the chain be a good approximation to the posterior, and that SMC meta-inference is sufficiently accurate. For some problems, this may be less expensive than the long chain required by AIS.

\subsection{Agglomerative Clustering for Dirichlet Process Mixtures}
A promising application of RAVI is to transform heuristic randomized algorithms into 
unbiased and consistent Monte Carlo estimators, by using them as proposal distributions. In this section, we design a RAVI inference strategy for clustering in Dirichlet process mixtures, based on a randomized agglomerative clustering algorithm (Inference Strategy~\hyperref[infstrat:agglom]{2}).
\begin{algorithm}[t]
    \label{infstrat:agglom}
    \SetAlgoLined\DontPrintSemicolon
    \scriptsize{
    \textbf{RAVI Inference Strategy 2:} Agglomerative Clustering\;
    \SetKwFunction{agglom}{agglom($X, K$).q}\SetKwFunction{agglomm}{agglom($X, K$).M($\Pi$).q}
    \SetKwProg{infalg}{Posterior Approx.}{}{}
    \SetKwInOut{Infers}{Target of inference}
    \SetKwInOut{Aux}{Auxiliary variables}
    \infalg{\agglom{}}{
    \Infers{partition $\Pi$ of dataset $X$}
    \Aux{merge sequence $\textit{mv}_{1:|X|-|\Pi|}$}
    \nl $\Pi \gets \{\{x\} \mid x \in X\}$ \tcp*{Initial partition}
    \nl \For{$l \in 1, \dots, |X|$}{
        \nl \For{\text{unordered pair }$\{i, j\}$\text{ of clusters in }$\Pi$}{
            \nl $w_{\{i,j\}} \gets \pi((\Pi \setminus \{i, j\}) \cup \{i \cup j\})$\;
        }
        \nl $w_{\text{stop}} = \pi(\Pi)$\;
        \nl $\textit{All} \gets \{\text{stop}\} \cup \{\{i, j\} \mid i, j \in \Pi\}$\;
        \nl $\textit{mv}_l \sim \text{Discrete}(\{m \Rightarrow w_m \mid m \in \textit{All}\})$ \;
        \nl \lIf{$\textit{mv}_l = \text{stop}$}{$\textbf{break}$}
        \nl $\Pi \gets (\Pi \setminus \textit{mv}_l) \cup (\cup \textit{mv}_l)$\tcp*{Perform the merge}
    }
    \nl \Return{$\Pi$}\;}{}
    \setcounter{AlgoLine}{0}
    \SetKwProg{metaalg}{Meta-Posterior Approx.}{}{}
    \metaalg{\agglomm{}}{
    \Infers{merge sequence $\textit{mv}_{1:|X|-|\Pi|}$}
    \Aux{particles $\textit{mv}_{1:|X|-|\Pi|}^{1:K}$, ancestors $a_{1:|X|-|\Pi|}^{1:K}$}
    \nl \lFor{$k \in 1,\dots, K$}{$\Pi_0^k, \textit{tr}_k  \gets \{\{x\} \mid x \in X\}, []$}
    \nl \For{$l \in 1, \dots, |X|-|\Pi|$}{
        \nl \For{$k \in 1,\dots,K$}{
            \nl \For{\text{unordered pair }$\{i, j\}$\text{ in }$\Pi_{l-1}^k$}{
            \nl $w_{\{i,j\}} \gets \pi((\Pi_{l-1}^k \setminus \{i, j\}) \cup \{i \cup j\})$\;
        }
        \nl $w_{\text{stop}} = \pi(\Pi_{l-1}^k)$\;
        \nl $\textit{All} \gets \{\text{stop}\} \cup \{\{i, j\} \mid i, j \in \Pi_{l-1}^k\}$\;
        \nl $\textit{Ok} \gets \{\{i, j\} \in \textit{All} \mid \exists c \in \Pi. i \cup j \subseteq c\}$\;
        \nl $\textit{mv}_l \sim \text{Discrete}(\{m \Rightarrow w_m \mid m \in \textit{Ok}\})$ \;
        \nl $\Pi_{l-1}^k \gets (\Pi_{l-1}^k \setminus \textit{mv}_l) \cup (\cup \textit{mv}_l)$\;
        \nl $\textit{tr}_k \gets [\textit{tr}_k\dots, \textit{mv}_l]$\;
        \nl $W_l^k \gets \frac{\sum_{m \in \textit{Ok}} w_m}{\sum_{m \in \textit{All}} w_m}$\;
        }

        \nl \For{$k \in 1,\dots,K$}{
            \nl $a_l^k \sim \text{Discrete}(W_l^{1:K})$ 
            \tcp*{resampling step}
            \nl $\Pi_l^k, \textit{tr}_k \gets \Pi_{l-1}^{a_l^k}, \textit{tr}_{a_l^k}$\;
        }
    }
    \nl \Return{$[\textit{tr}_1\dots, \text{stop}]$}}
    \SetKwProg{mmetaalg}{Meta-Meta-Posterior }{}{}
    \SetKwFunction{agglommm}{agglom($X, K$).M($\Pi$).M($\textit{mv}_{1:|X|-|\Pi|}$).q}
    \mmetaalg{\agglommm{}}{
        \tcp{Conditional SMC (omitted for space, but similar to that of \texttt{rmcvi})}
    }
}
\vspace{-5mm}
\end{algorithm} 

\textbf{Datasets and Models.} We test our algorithm on three clustering problems. The first is a synthetic 1D dataset sampled from a Dirichlet process (DP) mixture prior. The second is a standard benchmark dataset of galaxy velocities~\citep{drinkwater2004large,cusumano2017aide}, which we model using a DP mixture with Gaussian likelihoods and $\alpha = 1$.  The last is a data-cleaning task, correcting typos in 1k strings from Medicare records~\citep{MedicareHosp}. We adapt the generative model of~\citet{lew2021pclean}. Using an English character-level bigram model $H(s) = h(s_1) \prod_{i=2}^{|s|} h(s_i \mid s_{i-1})$, we model the data $\{y_i\}$  with a DP prior:
$$G \sim DP(H, \alpha = 1.0), \quad  x_i \mid G \sim G, \quad y_i \mid x_i \sim f(\cdot \mid x_i)$$
Here, the likelihood $f(y_i \mid x_i)$ models typos. We set $f$ to be
\[
f(y_i \mid x_i) \propto 
 \begin{cases} 
      \mathbf{1}[x_i = y_i] & (x_i, y_i) \not\in \mathcal{L} \times \mathcal{L} \\
      \frac{\text{NegBin}(\tau(x_i, y_i); \lceil \frac{|s|}{5} \rceil, 0.9)}{(5.09|s|)^{\tau(x_i, y_i)}} & (x_i, y_i) \in \mathcal{L} \times \mathcal{L}
  \end{cases},
\]
where $\tau(x_i, y_i)$ is the Damerau-Levenshtein edit distance between $x_i$ and $y_i$, and $\mathcal{L}$ is the set of all observed strings $\{y \mid \exists i. \, y = y_i\}$.
\footnote{We assume that the data $\mathcal{L}$ includes at least one example of every clean string. 
When $x_i \in \mathcal{L}$, we model a negative-binomially distributed number of typos, where the number of trials depends on the length of the string.}
We perform inference in a collapsed version of the model, with the $x_i$ marginalized out:
\begin{align*}
    \Pi &\sim CRP(n = N, \alpha = 1.0)\\
    y_I \mid \Pi &\sim F(y_I).
\end{align*}
Here $\Pi$ is a partition, $I$ ranges over the components of $\Pi$ (each of which is a subset of data indices), and $F(y_I) = \sum_{x \in \mathcal{L}} h(x) \prod_{i \in I} f(y_i \mid x)$ is the marginal likelihood of $y_I$ as a sequence of noisy observations of a latent string.


\textbf{Baseline.} We compare to an SMC baseline, inspired by PClean's inference~\citep{lew2021pclean}, that targets a sequence of posteriors, where the $t^{\text{th}}$ posterior incorporates the first $t$ datapoints. 
The SMC proposal is locally optimal, assigning the newest datapoint to an existing component $I$ with probability proportional to $\frac{|I|}{t + \alpha - 1} \cdot F(y_{I} \cup \{y_t\})$, or to a new component with probability proportional to $\frac{\alpha}{t + \alpha - 1} \cdot F(\{y_t\})$. We do not compare to a Gibbs sampling baseline, as Gibbs sampling does not yield marginal likelihood estimates, but do perform a Gibbs rejuvenation sweep every 20 iterations of SMC. 

\textbf{RAVI algorithm.} We apply Algorithm~\hyperref[alg:alg1]{1} to the inference strategy $\texttt{agglom}$ (Inference Strategy~\hyperref[infstrat:agglom]{2}). The strategy is based 
on a randomized agglomerative clustering algorithm: each datapoint begins 
in its own cluster (L1), and we repeatedly 
choose to either merge two clusters (L9) or stop and propose the current partition (L8). The sequence of merge 
decisions $\textit{mv}_1, \dots, \textit{mv}_{|X|-|\Pi|}$ are the auxiliary variables 
of our proposal distribution; the final output is the proposed clustering $\Pi$. Our meta-inference $\texttt{agglom}(X, K).\mathcal{M}(\Pi).q$ infers the sequence of merges from the observed clustering $\Pi$, using $K$-particle SMC with proposals that mimic the forward process but choose only from a restricted set $\textit{Ok}$ of possible merges (L8), to avoid making any choices that disagree with $\Pi$. SMC introduces additional auxiliary variables, so we also include a conditional SMC meta-meta-posterior approximation (not shown, but nearly identical to~\hyperref[infstrat:mcvi]{\texttt{rmcvi}}'s).

\textbf{Results.} Table~\ref{tab:experiments} shows 
average log marginal likelihood estimates; higher is 
better. On synthetic Gaussian data, the algorithms 
perform comparably. On the galaxy data, RAVI
agglomerative clustering finds modes that SMC misses,
leading to a 3-nat improvement in the 
average log marginal likelihood. In the Medicare 
data example, SMC misses the ground-truth clustering  
and hypothesizes many unlikely typos to 
explain the data. The RAVI agglomerative clustering is
less greedy, considering $O(N^2)$ possible merges at each 
step, rather than $O(N)$. As such, it is able to 
find the ground truth clustering, correctly identifying 
all typos (unlike PClean~\citep{lew2021pclean}, which achieves 
only 90\% accuracy on this dataset) and reporting a log marginal likelihood thousands of 
nats higher than the SMC algorithm.

\begin{table}[]
\scriptsize{
\begin{tabular}{ll|l}
& & $\hat{\mathcal{L}}$ \\ 
\hline
\multicolumn{2}{l|}{Gaussian likelihood~\citep{cusumano2017aide}, synthetic data}     &   \\ 
        & SMC + adapted proposals  &   $-125.09 \pm 0.38 $                   \\
        & RAVI agglomerative clustering                   &  ${-125.97 \pm 1.62}$        \\ 
\multicolumn{2}{l|}{Gaussian likelihood~\citep{cusumano2017aide}, Galaxy data~\citep{drinkwater2004large}}     &   \\ 
        & SMC + adapted proposals                &   $-426.20 \pm 1.26  $                   \\
        & RAVI agglomerative clustering                   &  $\mathbf{-423.03 \pm 0.94}$                     \\ 
\multicolumn{2}{l|}{PClean typos likelihood~\citep{lew2021pclean}, Hospital data~\citep{MedicareHosp}}&                       \\
        & SMC + adpated proposals       &      $-40,239 \pm 1,532$ \\
        & RAVI agglomerative clustering                 &      $\mathbf{-13,851.0 \pm 0.01}  $      \\ 
\end{tabular}
}
\caption{RAVI agglomerative clustering vs. SMC baseline.}
\label{tab:experiments}
\vspace{-6mm}
\end{table}

\vspace{-3mm}
\section{RELATED WORK AND DISCUSSION}

\textbf{Related work.} RAVI builds on and generalizes 
recent work from both the Monte Carlo and variational 
inference literatures. For example, \citet{salimans2015markov} and \citet{ranganath2016hierarchical} showed how 
auxiliary variables could be used to 
construct and optimize variational bounds for 
specific families of expressive variational approximations.
\citet{sobolev2019importance} presented tighter bounds 
in a more general setting. RAVI is a further 
generalization, in two directions: first, we show 
that these bounds arise from particular 
choices of meta-inference strategy, and can be 
tightened by improving meta-inference; and second, we
extend the results to the Monte Carlo setting, enabling 
learned variational families to be used as IS, SMC, or MH proposals. We also provide general theorems about the 
variance of RAVI samplers and the bias of RAVI variational bounds, which can be applied to analyze both new and existing algorithms.

RAVI is also related to other compositional or unifying frameworks for thinking about broad classes of inference algorithms~\citep{mansinghka2014venture,zinkov2016composing,scibior2018denotational,scibior2018functional,mansinghka2018probabilistic,cusumano2019gen, stites2021learning, neklyudov2020involutive, zimmermann2021nested, andrieu2020general, storvik2011flexibility, finke2015extended, finke2019importance}, some of which involve recursive constructions~\citep{naesseth2015nested, del2006sequential, domke2019divide}. However, to our knowledge, RAVI's inference strategies are novel. For example, although RAVI and Nested IS (NIS)~\citep{naesseth2015nested} are both approaches to inference with `intractable proposals,' NIS \textit{approximately samples} a proposal distribution with a \textit{tractable} (unnormalized) density, whereas RAVI \textit{approximates the density} of a proposal that \textit{can} be simulated tractably, but whose marginal density (even unnormalized) is intractable.
As another example, \citet{domke2019divide}'s framework of {\it estimator-coupling pairs} constructs variational bounds and marginal likelihood estimators recursively, but unlike in RAVI, 
the posterior approximations cannot be used to formulate objectives for \textit{amortized} VI or as components of Metropolis-Hastings proposals.

Finally, researchers have used meta-inference to construct bounds on KL divergences~\citep{cusumano2017aide} and other information-theoretic quantities~\citep{saad2022estimators}. In Appendix~\ref{sec:other-applications}, we show how to apply such bounds in the general RAVI setting.

\textbf{Outlook and Limitations.} RAVI expands the design space for Monte Carlo and variational inference. It gives unifying correctness proofs for over a dozen methods from the literature, and novel theorems that  characterize their behavior. Experiments show that RAVI helps to design algorithms that significantly improve accuracy over previously introduced Monte Carlo and variational inference methods.
However, some difficulties remain. For example, the gradient estimators we present (Algs.~\hyperref[alg:alg3]{3} and~\hyperref[alg:alg4]{4})
have high variance for some strategies $\mathcal{S}$; in Appendix~\ref{sec:reparam}, we give estimators that exploit the reparameterization trick, but they only help when the proposals in $\mathcal{S}$ can be 
reparameterized, which is not the case, e.g., for SMC. In these cases, RAVI can still be used to derive 
objectives for optimization, but practitioners will need other ways of reducing the variance of gradient estimates; many results from 
the literature~\citep{mnih2016variational,tucker2018doubly} should apply. 

Another difficulty is that RAVI algorithms can be complex to implement. We are exploring an automated implementation based on probabilistic programming languages~\citep{cusumano2019gen,van2018introduction}: if the posterior and meta-posterior approximations in a RAVI strategy $\mathcal{S}$ are given as probabilistic programs, we can provide Algs. 1-5 as higher-order functions, which automate the necessary densities, gradients, and MCMC acceptance probabilities. This could be viewed as a generalization of existing 
PPL support for \textit{programmable inference}~\citep{mansinghka2014venture,mansinghka2018probabilistic,cusumano2019gen,lew2019trace}. 

\begin{acknowledgements} 
    The authors are grateful to Feras Saad, Tan Zhi-Xuan, Ben Sherman,
    Cameron Freer, George Matheos, Sam Witty, McCoy Becker, Jan-Willem van de Meent, 
    Sam Stites, Eli Sennesh, Cathy Wong, and Nishad Gothoskar for useful conversations and feedback, and to our anonymous referees for helpful feedback on
    earlier drafts of the paper.
    This material is based on work supported by the NSF Graduate Research Fellowship under Grant No. 1745302.
\end{acknowledgements}

\bibliography{lew_657}

\onecolumn

\appendix

\section*{Supplementary Material for ``Recursive Monte Carlo and Variational Inference with Auxiliary Variables''}

This document and the accompanying code files contain supplementary material for the submission ``Recursive Monte Carlo and Variational Inference with Auxiliary Variables.'' In particular, we provide:

\begin{enumerate}
    \item In Section~\ref{sec:proofs}, \textbf{proofs} of Theorems 1-4.
    
    
    \item In Section~\ref{sec:appendix-examples}, \textbf{RAVI inference strategies for many existing algorithms}.
    
    \item In Section~\ref{sec:even-odd}, a further discussion of the {\bf absolute continuity requirements} for RAVI and how they can be relaxed.
    
    \item In Section~\ref{sec:other-applications}, \textbf{other applications} of RAVI inference strategies, to parameterize rejection sampling and KL divergence estimation algorithms.
\end{enumerate}

\section{Omitted Proofs.}
\label{sec:proofs}

Throughout this section, we use the notation introduced in Section~\ref{sec:theory}: the random variable $\hat{Z}(\tilde{\pi}, \mathcal{S})$ is the weight returned by $\texttt{IMPORTANCE}(\tilde{\pi}, \mathcal{S})$, and $\check{Z}(\tilde{\pi}, \mathcal{S})$ is the reciprocal of the weight returned by $\texttt{HME}(\tilde{\pi}, x, \mathcal{S})$, for $x \sim \pi$.

\subsection{Proof of Theorem 1.}

\textbf{Theorem 1.} \textit{
    Let $\tilde{\pi}(x) = Z\pi(x)$ be an unnormalized target density, and $\mathcal{S}$ an inference strategy targeting $\pi(x)$. Then:
    \begin{itemize}
        \item $\texttt{IMPORTANCE}(\mathcal{S}, \tilde{\pi})$ generates $(x, \hat{Z})$ with $x \sim \mathcal{S}.q$ and $\mathbb{E}[\hat{Z} \mid x] = Z\frac{\pi(x)}{\mathcal{S}.q(x)}$. Furthermore, the unconditional expectation $\mathbb{E}[\hat{Z}(\tilde{\pi}, \mathcal{S})] = Z$.
        \item $\mathbb{E}[{\check{Z}}(\tilde{\pi}, \mathcal{S})^{-1}] = \mathbb{E}_{x \sim \pi}[\texttt{HME}(\mathcal{S}, x, \tilde{\pi})] = Z^{-1}.$
    \end{itemize}
}

\textbf{Proof.} The proof is by induction on the level of nesting 
present in the inference strategy. 

First consider the case where $\mathcal{S}.q$ has a tractable 
marginal density. Then:
\begin{itemize}
    \item \texttt{IMPORTANCE} samples $x \sim \mathcal{S}.q$ on line 2, and 
computes $\hat{Z} = \frac{\tilde{\pi}(x)}{\mathcal{S}.q(x)} = Z\frac{\pi(x)}{\mathcal{S}.q(x)}$ exactly (lines 3 and 7). By the standard importance sampling argument, the unconditional expectation $\mathbb{E}[\hat{Z}(\tilde{\pi}, \mathcal{S})] = \mathbb{E}_{x \sim \mathcal{S}.q}[Z\frac{\pi(x)}{\mathcal{S}.q(x)}]=Z\mathbb{E}_{x \sim \pi}[1] = Z$. (This argument relies on the fact that, because $\mathcal{S}$ targets $\pi$, $\pi$ is absolutely continuous with respect to $\mathcal{S}.q$.)
    \item 
$\texttt{HME}(\mathcal{S}, x, \tilde{\pi})$ returns exactly $\frac{\mathcal{S}.q(x)}{\tilde{\pi}(x)}$ (lines 2 and 5), and $$\mathbb{E}_{x \sim \pi}\left[\frac{\mathcal{S}.q(x)}{\tilde{\pi}(x)}\right] = \int \pi(x) \frac{\mathcal{S}.q(x)}{Z\pi(x)}\text{d}x = \frac{1}{Z} \int \mathcal{S}.q(x) \text{d}x = \frac{1}{Z},$$ where the last step follows because $\mathcal{S}.q$ is a normalized probability density, and $\mathcal{S}.q$ is absolutely continuous with respect to $\pi$.
\end{itemize}

Now consider the inductive step. Assume $\mathcal{S}.q(x) = \int \mathcal{S}.q(r, x)\text{d}r$ and that for all $x$, the theorem holds for the inference strategy $\mathcal{S}.\mathcal{M}(x)$ and the unnormalized target distribution $\mathcal{S}.q(\cdot, x)$. In this case:

\begin{itemize}
    \item On line 5, \texttt{IMPORTANCE} generates $x \sim \mathcal{S}.q$ and $r \sim \mathcal{S}.q(r \mid x)$. In the call to $\texttt{HME}$, the unnormalized target distribution is $\mathcal{S}.q(\cdot, x)$, and so the normalizing constant is $\mathcal{S}.q(x)$ and the normalized target is $\mathcal{S}.q(r \mid x)$. By the inductive hypothesis, the call to $\texttt{HME}$ on line 6 returns an unbiased estimate of the normalizing constant's reciprocal, i.e. $\mathbb{E}[w \mid x] = \frac{1}{\mathcal{S}.q(x)}$. Since \texttt{IMPORTANCE} returns $\hat{Z} = w\tilde{\pi}(x)$ on line 7, this implies that $\mathbb{E}[\hat{Z} \mid x] = \frac{\tilde{\pi}(x)}{\mathcal{S}.q(x)} = Z\frac{\pi(x)}{\mathcal{S}.q(x)}$. From this, the same standard importance sampling argument as above shows that the unconditional expectation $\mathbb{E}[\hat{Z}(\tilde{\pi}, \mathcal{S})] = Z$.
    \item On line 4, \texttt{HME} calls \texttt{IMPORTANCE} on the 
    unnormalized target $\mathcal{S}.q(\cdot, x)$, and so by the inductive hypothesis, $\mathbb{E}[w] = \mathcal{S}.q(x)$ (the normalizing constant). On line 5, the returned weight has expectation $\mathbb{E}_{x \sim \pi}\left[\frac{w}{\tilde{\pi}(x)}\right]=\frac{1}{Z} \int \pi(x) \cdot \frac{\mathcal{S}.q(x)}{\pi(x)} \text{d}x = \frac{1}{Z}$, where the last equality again follows because $\mathcal{S}.q$ is a normalized density, and $\mathcal{S}.q$ is absolutely continuous with respect to $\pi$.
\end{itemize}

\subsection{Proof of Theorem 2}

\begin{lemma}
    For an inference strategy $\mathcal{S}$ targeting $p(x \mid y)$, if $\mathcal{S}.q(x) = \int \mathcal{S}.q(r, x)\text{d}r$ has an intractable marginal density, then:
    \[
    \LL(p, y, \mathcal{S}) = \E_{x \sim \mathcal{S}.q} [\log p(x, y) - \UU(\mathcal{S}.q, x, \mathcal{S}.\mathcal{M}(x)]
    \]
    and
    \[
    \UU(p, y, \mathcal{S}) = \E_{x \sim p( \cdot | y) } [ \log p(x, y) - \LL(\mathcal{S}.q, x, \mathcal{S}.\mathcal{M}(x)) ]
    \]
\end{lemma}

\textbf{Proof.} For the first conclusion,

\begin{align}
\LL(p, y, \mathcal{S})
    &= \E[\log \hat{Z}(p(\cdot \mid y), \mathcal{S})]\\
    &= \E\left[\log \frac{p(x, y)}{\check{Z}(\mathcal{S}.q(\cdot, x), \mathcal{S}.\mathcal{M}(x))}\right]\\
    &= \E_{x \sim \mathcal{S}.q}[\E[\log p(x, y) - \log \check{Z}(\mathcal{S}.q(\cdot, x), \mathcal{S}.\mathcal{M}(x)) \mid x]]\\
    &= \E_{x \sim \mathcal{S}.q}[\log p(x, y) - \mathbb{E}[\log \check{Z}(\mathcal{S}.q(\cdot, x), \mathcal{S}.\mathcal{M}(x)) \mid x]]\\
    &= \E_{x \sim \mathcal{S}.q} [\log p(x, y) - \UU(\mathcal{S}.q, x, \mathcal{S}.\mathcal{M}(x)) ]
\end{align}
The same approach, but with $\E [ \log \check{Z} ]$, can be used to prove the other conclusion.

\noindent\textbf{Theorem 2.} \textit{
Given a model $p_\theta(x, y)$ and an inference strategy $\mathcal{S}_\theta$ targeting $p_\theta(x \mid y)$,
Alg. 3 yields unbiased estimates of $\LL(p, y, \mathcal{S})$ and of $\grad \LL(p, y, \mathcal{S})$.
Furthermore, when $(x, y) \sim p_\theta$, Alg. 4 yields
(i) $\hat{U}$ such that
$\E [ \hat{U} \mid y ] = \UU(p, y, \mathcal{S})$,
(ii) $\widehat{\grad}$ such that
$\E [ \widehat{\grad} ] = \grad \mathbb{E}_{y \sim p_\theta}[\UU(p, y, \mathcal{S})]$,
and (iii) a value $\mathbf{g}$ such that for any function $R$ that does not depend on $\theta$,
$\E [ \mathbf{g} \cdot R(y) ] = \grad \E_{y \sim p_\theta} [ R(y) ]$  if
$\grad \E_{y \sim p_\theta} [ R(y) ]$ is defined.}

\textbf{Proof.} The proof is by induction on the level of nesting present in the inference strategy.

First consider inference strategies $\mathcal{S}$ with tractable proposals $\mathcal{S}.q(x)$. In this case $\texttt{ELBO}\nabla$ generates $x \sim \mathcal{S}.q$ and returns $\hat{L} = \log p(x, y) - \log \mathcal{S}.q(x)$ and $\widehat{\grad} = \grad (\log p(x, y) - \log \mathcal{S}.q(x)) + (\grad \log \mathcal{S}.q(x))(\log p(x, y) - \log \mathcal{S}.q(x))$. Clearly, $\E_{x \sim \mathcal{S}.q}[\hat{L}] = \E[\log \hat{Z}(p(\cdot, y), \mathcal{S})] = \LL(p, y, \mathcal{S})$. And by the log-derivative trick, $\E_{x \sim \mathcal{S}.q}[\widehat{\grad}] = \E[\grad (\log p(x, y) - \log \mathcal{S}.q(x))] = \E[\mathcal{L}(p, y, \mathcal{S})]$. When we apply $\texttt{EUBO}\nabla$ to $\mathcal{S}$ with $(x, y) \sim p$, it returns (1) $\hat{U} = \log p(x, y) - \log \mathcal{S}.q(x)$ (for which $\E[\hat{U} \mid y] = \UU(p, y, \mathcal{S})$), (2) $\widehat{\grad} = \grad (\log p(x, y) - \log \mathcal{S}.q(x)) + \grad \log p(x, y) (\log p(x, y) - \log \mathcal{S}.q(x))$ (for which, by the log-derivative trick, $\mathbb{E}[\widehat{\grad}] = \grad \E_{y \sim p}[\UU(p, y, \mathcal{S})]$), and (3) $\mathbf{g} = \grad \log p(x, y)$. This last return value satisfies the spec for $\mathbf{g}$ because if $R$ does not depend on $\theta$, then $\mathbb{E}_{(x, y) \sim p}[R(y) \cdot \grad\log p(x, y)] = \int\int p(x, y) \cdot 
\frac{\grad p(x, y)}{p(x, y)} \cdot R(y) \text{d}x \text{d}y = \grad \int \int p(x, y) R(y) \text{d}x \text{d}y = \grad \E [R(y)]$, as required. 

Now consider the inductive step. Assume the theorem holds for the inference strategy $\mathcal{S}.\mathcal{M}(x)$ and joint distribution $\mathcal{S}.q(r, x)$. 

We first consider $\texttt{ELBO}\nabla$. It generates $(r, x) \sim \mathcal{S}.q$ before calling $\texttt{EUBO}\nabla$, which by induction returns $(\hat{U}, \widehat{\grad}, \mathbf{g})$ such that:

\begin{enumerate}
    \item $\mathbb{E}[\hat{U} \mid x] = \UU(\mathcal{S}.q, x, \mathcal{S}.\mathcal{M}(x))$
    \item $\E[\widehat{\grad}] = \grad\E_{x \sim \mathcal{S}.q}[\UU(\mathcal{S}.q, x, \mathcal{S}.\mathcal{M}(x))]$
    \item $\E[g \cdot R(x)] = \grad \E_{x \sim \mathcal{S}.q}[R(x)]$ for all valid $R$.
\end{enumerate}

$\texttt{ELBO}\nabla$ computes its first return value, $\hat{L}$, as $\log p(x, y) - \hat{U}$, so
\begin{align*}
    \E[\hat{L}] &= \E[\log p(x, y) - \hat{U}]\\ 
    &= \E_{x \sim \mathcal{S}.q}[\E[\log p(x, y) - \hat{U} \mid x]]\\
    &= \E_{x \sim \mathcal{S}.q}[\log p(x, y) - \E[\hat{U} \mid x]]\\
    &= \E_{x \sim \mathcal{S}.q}[\log p(x, y)-\UU(\mathcal{S}.q, x, \mathcal{S}.\mathcal{M}(x))]\\ 
    &= \LL(p, y, \mathcal{S}),
\end{align*}
where the fourth equality holds by the inductive hypothesis and the final one by Lemma 1. Its second return value is computed as $\widehat{\grad}' = \grad\log p(x, y) + \mathbf{g}\log p(x, y) - \widehat{\grad}$, and so
\begin{align*}
    \E [ \widehat{\grad}' ]
        &= \E \left[ \grad \log p(x, y) + \mathbf{g} \cdot \log p(x, y) - \widehat{\grad} \right]\\
        &= \E \left[ \grad \log p(x, y) \right]
            + \grad \E_{x \sim \mathcal{S}.q} [ \log \boxed{p}(x, y) ]
            - \grad \E_{x \sim \mathcal{S}.q} [ \mathcal{U}(\mathcal{S}.q, x, \mathcal{S}.\mathcal{M}(x)) ]\\
        &= \grad \E \left[ \log p(x, y) \right]
            - \grad \E_{x \sim \mathcal{S}.q} [ \mathcal{U}(\mathcal{S}.q, x, \mathcal{S}.\mathcal{M}(x)) ]\\
        &= \grad \E_{x \sim \mathcal{S}.q} [ \log p(x, y) - \UU(\mathcal{S}.q, x, \mathcal{S}.\mathcal{M}(x)) ]\\
        &= \grad \LL(p, y, \mathcal{S}),
\end{align*}
where $\boxed{p}(x, y)$ denotes the distribution $p(x, y)$ but without a dependence on $\theta$, for the purposes of differentiation with respect to $\theta$. The second equality holds by the inductive hypothesis about $\mathbf{g}$ (with $R(x) = \log \boxed{p}(x, y)$) and about $\widehat{\grad}$, and the third uses the log-derivative trick. The final equation is due to Lemma 1.

We now turn to $\texttt{EUBO}\nabla$. 
By induction, the call to $\texttt{ELBO}\nabla$ satisfies the theorem, and so: \begin{enumerate}
    \item $\E [ \hat{L} ] = \LL(\mathcal{S}.q, x, \mathcal{S}.\mathcal{M}(x))$
    \item $\E [ \widehat{\grad} ] = \grad\LL(\mathcal{S}.q, x, \mathcal{S}.\mathcal{M}(x))$
\end{enumerate}

We treat each of the return values, $(\hat{U}, \widehat{\grad}, \mathbf{g})$, in sequence. We view them as random variables, accounting for stochasticity in the algorithm as well as the inputs $(x, y)$, which are assumed in the theorem's statement to be jointly distributed according to $p$.

First, $\hat{U}$ is computed as $\log p(x, y) - \hat{L}$, and so 
\begin{align*}
\E [ \hat{U} | y ] 
    &= \E_{x \sim p(\cdot | y)} [ \E [ \log p(x, y) - \hat{L} | x, y ] ]\\
    &= \E_{x \sim p(\cdot | y) } [ \log p(x, y) - \LL(\mathcal{S}.q, x, \mathcal{S}.\mathcal{M}(x)) ]\\
    &= \UU(p, y, \mathcal{S}).
\end{align*}
Next, $\E[ \widehat{\grad}']$:
\begin{align*}
    \E[\widehat{\grad}']
    &= \E_{x, y \sim p} \left[
            \E \left[
                \grad \log p(x, y)
                + (\grad \log p(x, y)) \cdot \hat{U}
                - \widehat{\grad} | x, y
            \right]
        \right]\\
    &= \E_{x, y \sim p} \left[
            \grad \log p(x, y) +
            (\grad \log p(x, y))
                \cdot \E \left[ \hat{U} | x, y \right]
            - \E \left[\widehat{\grad} | x, y \right]
        \right]\\
    &= \E_{x, y \sim p} \left[
        \grad \log p(x, y) +
        (\grad \log p(x, y))
            \cdot \E \left[ \log p(x, y) - \hat{L} \mid x, y \right]
        - \grad \LL(\mathcal{S}.q, x, \mathcal{S}.\mathcal{M}(x))
    \right]\\
    &= \E_{x, y \sim p} \left[
        \grad \log p(x, y) 
        - \grad \LL(\mathcal{S}.q, x, \mathcal{S}.\mathcal{M}(x))
        + (\grad \log p(x, y))
            \cdot (\log p(x, y) - \LL(\mathcal{S}.q, x, \mathcal{S}.\mathcal{M}(x))
    \right]\\
    &= \grad \E_{x, y \sim p} \left[
            \log p(x, y) - \LL(\mathcal{S}.q, x, \mathcal{S}.\mathcal{M}(x))
        \right]\\        
    &= \grad \E_{y \sim p} [ \E_{x \sim p(\cdot | y) } [ \log p(x, y) - \LL(\mathcal{S}.q, x, \mathcal{S}.\mathcal{M}(x)) ] ]\\
    &= \grad \E_{y \sim p} [ \UU(p, y, \mathcal{S}) ].    
\end{align*}
Finally, we consider $\E_{y \sim p} [ \E [ \mathbf{g} \cdot R(y) | y ] ]$ (and recall that $R(y)$ is not to be treated as a function of $\theta$):
\begin{align*}
\E_{y \sim p} [ \E [ \mathbf{g} \cdot R(y) | y ] ]
    &= \E_{y \sim p} \left[ \E [ (\grad \log p(x, y)) \cdot R(y) | y ] \right]\\
    &= \E_{x, y \sim p} \left[ (\grad \log p(x, y)) \cdot R(y) \right]\\
    &= \grad \E_{x, y \sim p} \left[ R(y) \right]\\
    &= \grad \E_{y \sim p} [ R(y) ].
\end{align*}

\subsection{Proof of Theorem 3}

\noindent\textbf{Theorem 3.} {\it Consider an unnormalized target distribution $\tilde{\pi}(x) = Z\pi(x)$ and an inference strategy $\mathcal{S}$ targeting $\pi(x)$. Then the relative variances of the estimators $\hat{Z}(\tilde{\pi}, \mathcal{S})$ and $\check{Z}(\tilde{\pi}, \mathcal{S})$ are given by the following recursive equations:}
\begin{align*}
\text{Var}_{\hat{Z}}&(\pi, \mathcal{S}) = \chi^2(\pi || \mathcal{S}.q) \, +\\
 & \mathbb{E}_{x \sim \mathcal{S}.q}\left[\left(\frac{\pi(x)^2}{\mathcal{S}.q(x)^2}\right) \cdot \text{Var}_{\check{Z}}(\mathcal{S}.q(\cdot \mid x), \mathcal{S}.\mathcal{M}(x))\right]\\
\text{Var}_{\check{Z}}&(\pi, \mathcal{S}) = \chi^2(\mathcal{S}.q || \pi) + \\
& \mathbb{E}_{x \sim \pi}\left[\left(\frac{\mathcal{S}.q(x)^2}{\pi(x)^2}\right) \cdot \text{Var}_{\hat{Z}}(\mathcal{S}.q(\cdot \mid x), \mathcal{S}.\mathcal{M}(x))\right]
%
\end{align*}
{\it When $\mathcal{S}.q$ is tractable, the second term of each sum is 0.}

\textbf{Proof.} The proof is by induction on the level of nesting present in the inference strategy $\mathcal{S}$. 

First suppose $\mathcal{S}.q$ has a tractable marginal density. Then:

\begin{itemize}
    \item $\hat{Z}(\pi, \mathcal{S})$ is the normalized importance weight 
    $\frac{\pi(x)}{\mathcal{S}.q(x)}$, with $x \sim \mathcal{S}.q$. So the relative variance is:
    $$\text{Var}_{\hat{Z}}({\pi}, \mathcal{S}) = \text{Var}\left(\hat{Z}({\pi}, \mathcal{S})\right) = \mathbb{E}_{x \sim \mathcal{S}.q}\left[\frac{\pi(x)^2}{\mathcal{S}.q(x)^2}\right] - \mathbb{E}_{x \sim \mathcal{S}.q}\left[\frac{\pi(x)}{\mathcal{S}.q(x)}\right]^2 = \mathbb{E}_{x \sim \mathcal{S}.q}\left[\frac{\pi(x)^2}{\mathcal{S}.q(x)^2} - 1\right] = \chi^2(\pi || \mathcal{S}.q),$$
    where the third equality holds because $\pi$ is a normalized density and $\pi$ is absolutely continuous with respect to $\mathcal{S}.q$.

    \item $\check{Z}({\pi}, \mathcal{S})$ is the weight $\frac{\pi(x)}{\mathcal{S}.q(x)}$, with $x \sim \pi$. Then the relative variance $$\text{Var}_{\check{Z}}({\pi}, \mathcal{S}) = \text{Var}\left(\check{Z}(\pi, \mathcal{S})^{-1}\right) = \mathbb{E}_{x \sim \pi}\left[\frac{\mathcal{S}.q(x)^2}{\pi(x)^2}\right] - \mathbb{E}_{x \sim \pi}\left[\frac{\mathcal{S}.q(x)}{\pi(x)}\right]^2 = \mathbb{E}_{x \sim \pi}\left[\frac{\mathcal{S}.q(x)^2}{\pi(x)^2} - 1\right] = \chi^2(\mathcal{S}.q || \pi),$$ where the third equality holds because $\mathcal{S}.q$ is a normalized density and is absolutely continuous with respect to $\pi$.
\end{itemize}
Now consider the inductive step. Assume that for all $x$, the theorem holds of the strategy $\mathcal{S}.\mathcal{M}(x)$ targeting $\mathcal{S}.q(\cdot \mid x)$. $\text{Var}_{\hat{Z}}(\mathcal{S}.q(\cdot \mid x), \mathcal{S}.\mathcal{M}(x))$, for all $x$. Then:
\begin{itemize}
    \item The $\texttt{IMPORTANCE}(\pi, \mathcal{S})$ algorithm generates $x \sim \mathcal{S}.q$. It then calls \texttt{HME} (with $r \sim \mathcal{S}.q(\cdot \mid x)$) to obtain $w = \check{Z}(\mathcal{S}.q(\cdot, x), \mathcal{S}.\mathcal{M}(x))^{-1}$, and returns $\hat{Z} = w\pi(x)$. The variance of $\hat{Z}$ is then:
    \begin{align*}
    \text{Var}_{\hat{Z}}(\pi, \mathcal{S}) &= \text{Var}\left(\frac{\pi(x)}{\check{Z}(\mathcal{S}.q(\cdot, x), \mathcal{S}.\mathcal{M}(x))}\right)\\
    &= \mathbb{E}\left[\left(\frac{\pi(x)}{\check{Z}(\mathcal{S}.q(\cdot, x), \mathcal{S}.\mathcal{M}(x))}\right)^2 - 1\right]
    && (\mathbb{E}[\hat{Z}(\pi, \mathcal{S})]^2 = Z^2 = 1)\\
    &= \mathbb{E}\left[\left(\frac{\pi(x)}{\mathcal{S}.q(x)} \cdot \frac{\mathcal{S}.q(x)}{\check{Z}(\mathcal{S}.q(\cdot, x), \mathcal{S}.\mathcal{M}(x))}\right)^2 - 1\right]
    && \text{(divide and multiply by } \mathcal{S}.q(x))\\
    &= \mathbb{E}\left[\left(\frac{\pi(x)}{\mathcal{S}.q(x)} \cdot \frac{1}{\check{Z}(\mathcal{S}.q(\cdot \mid x), \mathcal{S}.\mathcal{M}(x))}\right)^2 - 1\right]
    &&(\mathcal{S}.q(x) \text{ is the normalizing constant of } \mathcal{S}.q(\cdot, x))\\
    &= \mathbb{E}\left[\left(\frac{\pi(x)}{\mathcal{S}.q(x)}\right)^2 \left(\mathbb{E}\left[{\check{Z}(\mathcal{S}.q(\cdot \mid x), \mathcal{S}.\mathcal{M}(x))}^{-2} \bigl\vert x\right]\right) - 1\right]\\
    &= \mathbb{E}\left[\left(\frac{\pi(x)}{\mathcal{S}.q(x)}\right)^2 \left(\text{Var}_{\check{Z}}(\mathcal{S}.q(\cdot \mid x), \mathcal{S}.\mathcal{M}(x)) + 1\right) - 1\right]
    && \text{(definition of Var}_{\check{Z}}(\cdot, \cdot))\\
    &= \mathbb{E}\left[\left(\frac{\pi(x)}{\mathcal{S}.q(x)}\right)^2 \left(\text{Var}_{\check{Z}}(\mathcal{S}.q(\cdot \mid x), \mathcal{S}.\mathcal{M}(x))\right) + \left(\frac{\pi(x)}{\mathcal{S}.q(x)}\right)^2 - 1\right]
    && \text{(distributing product over sum)}\\
    &= \mathbb{E}\left[\left(\frac{\pi(x)}{\mathcal{S}.q(x)}\right)^2 \left(\text{Var}_{\check{Z}}(\mathcal{S}.q(\cdot \mid x), \mathcal{S}.\mathcal{M}(x))\right)\right] + \chi^2(\pi || \mathcal{S}.q).\\
    \end{align*}

    \item The argument for $\check{Z}$ is largely the same:
    \begin{align*}
        \text{Var}_{\check{Z}}(\pi, \mathcal{S}) &= \text{Var}\left(\frac{\hat{Z}(\mathcal{S}.q(\cdot, x), \mathcal{S}.\mathcal{M}(x))}{\pi(x)}\right)\\
        &= \mathbb{E}\left[\left(\frac{\check{Z}(\mathcal{S}.q(\cdot, x), \mathcal{S}.\mathcal{M}(x))}{\pi(x)}\right)^2 - 1\right]
        && (\mathbb{E}[\check{Z}(\pi, \mathcal{S})^{-1}]^2 = Z^{-2} = 1)\\
        &= \mathbb{E}\left[\left(\frac{\mathcal{S}.q(x)}{\pi(x)} \cdot \frac{\hat{Z}(\mathcal{S}.q(\cdot, x), \mathcal{S}.\mathcal{M}(x))}{\mathcal{S}.q(x)}\right)^2 - 1\right]
        && \text{(divide and multiply by } \mathcal{S}.q(x))\\
        &= \mathbb{E}\left[\left(\frac{\mathcal{S}.q(x)}{\pi(x)} \cdot {\hat{Z}(\mathcal{S}.q(\cdot \mid x), \mathcal{S}.\mathcal{M}(x))}\right)^2 - 1\right]
        &&(\mathcal{S}.q(x) \text{ is the normalizing constant of } \mathcal{S}.q(\cdot, x))\\
        &= \mathbb{E}\left[\left(\frac{\mathcal{S}.q(x)}{\pi(x)}\right)^2 \left(\mathbb{E}\left[{\hat{Z}(\mathcal{S}.q(\cdot \mid x), \mathcal{S}.\mathcal{M}(x))}^{2} \bigl\vert x\right]\right) - 1\right]\\
        &= \mathbb{E}\left[\left(\frac{\mathcal{S}.q(x)}{\pi(x)}\right)^2 \left(\text{Var}_{\hat{Z}}(\mathcal{S}.q(\cdot \mid x), \mathcal{S}.\mathcal{M}(x)) + 1\right) - 1\right]
        && \text{(definition of Var}_{\hat{Z}}(\cdot, \cdot))\\
        &= \mathbb{E}\left[\left(\frac{\mathcal{S}.q(x)}{\pi(x)}\right)^2 \left(\text{Var}_{\hat{Z}}(\mathcal{S}.q(\cdot \mid x), \mathcal{S}.\mathcal{M}(x))\right) + \left(\frac{\mathcal{S}.q(x)}{\pi(x)}\right)^2 - 1\right]
        && \text{(distributing product over sum)}\\
        &= \mathbb{E}\left[\left(\frac{\mathcal{S}.q(x)}{\pi(x)}\right)^2 \left(\text{Var}_{\hat{Z}}(\mathcal{S}.q(\cdot \mid x), \mathcal{S}.\mathcal{M}(x))\right)\right] + \chi^2(\mathcal{S}.q || \pi).\\
    \end{align*}
\end{itemize}

\subsection{Proof of Theorem 4.}

\noindent\textbf{Theorem 4.}
{\it Consider a joint distribution $p(x, y)$ and an inference strategy $\mathcal{S}$ targeting $p(x \mid y)$. Then the following equations give the bias of ${\hat{\LL}}$ and ${\hat{\UU}}$ as estimators of $\log p(y)$:}
\begin{align*}
\text{Bias}_\mathcal{L}(p, y, \mathcal{S}) =&\, -\text{KL}(\mathcal{S}.q || p(\cdot \mid y)) \\ 
&-\mathbb{E}_{x \sim \mathcal{S}.q}[\text{Bias}_\mathcal{U}(\mathcal{S}.q, x, \mathcal{S}.\mathcal{M}(x))]\\
\text{Bias}_\mathcal{U}(p, y, \mathcal{S}) =&\, \text{KL}(p(\cdot \mid y) || \mathcal{S}.q)\\ 
&\,\,\,-\mathbb{E}_{x \sim p(\cdot \mid y)}[\text{Bias}_\mathcal{L}(\mathcal{S}.q, x, \mathcal{S}.\mathcal{M}(x))]
\end{align*}
{\it where the second term in each equation is 0 when $\mathcal{S}.q$ has a tractable marginal density.}

\textbf{Proof.}


In the base case, where $\mathcal{S}.q$ has a tractable marginal density, the theorem states that $\log p(y) - \LL(p, y, \mathcal{S}) = KL(\mathcal{S}.q || p(\cdot \mid y))$, the familiar relationship between the standard ELBO and the KL divergence. The $\UU$ case is similar:
\begin{align*}
    \text{Bias}_\UU(p, y, \mathcal{S}) 
    &= \E_{x \sim p(\cdot \mid y)}[\log p(x, y) - \log \mathcal{S}.q(x)] - \log p(y)\\
    &= \log p(y) + \E_{x \sim p(\cdot \mid y)}[\log p(x \mid y) - \log \mathcal{S}.q(x)] - \log p(y)\\ 
    &= KL(p(\cdot \mid y) || \mathcal{S}.q).
\end{align*}

Now consider the inductive step, in which $\mathcal{S}.q$ does not have a tractable marginal density. We assume the theorem holds for $\mathcal{S}.q$ and $\mathcal{S}.\mathcal{M}(x)$. Then: 
\begin{align*}
    \text{Bias}_\LL(p, y, \mathcal{S}) 
    &= \LL(p, y, \mathcal{S}) - \log p(y)\\
    &= \E_{x \sim \mathcal{S}.q}[\log p(x, y) - \UU(\mathcal{S}.q, x, \mathcal{S}.\mathcal{M}(x))] - \log p(y)\\
    &= \log p(y) + \E_{\sim \mathcal{S}.q}[\log p(x \mid y) - \UU(\mathcal{S}.q, x, \mathcal{S}.\mathcal{M}(x))]\\
    &= \E_{x \sim \mathcal{S}.q}[\log p(x \mid y) - \UU(\mathcal{S}.q, x, \mathcal{S}.\mathcal{M}(x))]\\
    &= \E_{x \sim \mathcal{S}.q}[\log p(x \mid y) - \log \mathcal{S}.q(x) + \log \mathcal{S}.q(x) - \UU(\mathcal{S}.q, x, \mathcal{S}.\mathcal{M}(x))]\\
    &= -KL(\mathcal{S}.q || p(\cdot \mid y)) + \E_{x \sim \mathcal{S}.q}[\log \mathcal{S}.q(x) - \UU(\mathcal{S}.q, x, \mathcal{S}.\mathcal{M}(x))]\\
    &= -KL(\mathcal{S}.q || p(\cdot \mid y)) - \E_{x \sim \mathcal{S}.q}[\text{Bias}_\UU(\mathcal{S}.q, x, \mathcal{S}.\mathcal{M}(x))].
\end{align*}

Nearly the same proof applies for $\UU$, flipping the necessary signs:
\begin{align*}
    \text{Bias}_\UU(p, y, \mathcal{S}) 
    &= \UU(p, y, \mathcal{S}) - \log p(y)\\
    &= \E_{x \sim p(\cdot \mid y)}[\log p(x, y) - \LL(\mathcal{S}.q, x, \mathcal{S}.\mathcal{M}(x))] - \log p(y)\\
    &= \log p(y) + \E_{\sim p(\cdot \mid y)}[\log p(x \mid y) - \LL(\mathcal{S}.q, x, \mathcal{S}.\mathcal{M}(x))]\\
    &= \E_{x \sim p(\cdot \mid y)}[\log p(x \mid y) - \LL(\mathcal{S}.q, x, \mathcal{S}.\mathcal{M}(x))]\\
    &= \E_{x \sim p(\cdot \mid y)}[\log p(x \mid y) - \log \mathcal{S}.q(x) + \log \mathcal{S}.q(x) - \LL(\mathcal{S}.q, x, \mathcal{S}.\mathcal{M}(x))]\\
    &= KL(p(\cdot \mid y) || \mathcal{S}.q) + \E_{x \sim p(x \mid y)}[\log \mathcal{S}.q(x) - \LL(\mathcal{S}.q, x, \mathcal{S}.\mathcal{M}(x))]\\
    &= KL(p(\cdot \mid y) || \mathcal{S}.q) - \E_{x \sim p(\cdot \mid y)}[\text{Bias}_\LL(\mathcal{S}.q, x, \mathcal{S}.\mathcal{M}(x))].
\end{align*}

\subsection{Stationarity of MCMC algorithm}
\label{sec:ravi-mcmc}
In Section~\ref{sec:inference-algs}, we mention that RAVI can be used to run Metropolis-Hastings kernels 
with proposals that have intractable densities. 
Here, we present and justify the algorithm.

Let $\tilde{\pi}(x) = \int \tilde{\pi}(r, x) \text{d}r = Z \int \pi(r, x) \text{d}r$ be a possibly unnormalized target density, and let $q(x'; x) = \int q(s, x'; x) \text{d}s$ be a proposal kernel mapping previous state $x$ to new state $x'$. We note that (1) both $\tilde{\pi}$ and $q$ have intractable marginal densities, and (2) the target marginal $\tilde{\pi}(x)$ itself may be unnormalized. As is typical in pseudomarginal MCMC, even this unnormalized target density cannot be evaluated pointwise, due to the additional nuisance variables $r$.

Now suppose we have a family of inference strategies $\mathcal{S}(x)$ targeting $\pi(r \mid x)$, and a family of inference strategies $\mathcal{M}(x, x')$ targeting $q(s \mid x'; x)$. Let $x$ be a starting position for our Markov chain. We can run Algorithm~\hyperref[alg:alg1]{1} on $\mathcal{S}$, targeting $\pi(r \mid x)$, to obtain an initial estimate $\hat{Z}_x$ of the unnormalized marginal density $\tilde{\pi}(x)$. Then Algorithm~\hyperref[alg:mh]{5} defines a stationary MCMC kernel for the target distribution $\pi(x)$, starting at input point $x$:

\begin{wrapfigure}{L}{0.5\textwidth}
    \begin{algorithm}[H]
    \SetAlgoLined\DontPrintSemicolon
    \label{alg:mh}
    \footnotesize{
    \textbf{Algorithm 5:} RAVI Metropolis-Hastings\;
    \KwIn{model $\tilde{\pi}(x) = Z \int \pi(r, x) \text{d}r$}
    \KwIn{proposal $q(x'; x) = \int q(s, x'; x) \text{d}s$}
    \KwIn{family $\mathcal{S}(x)$ of inference strategies targeting $\pi(r \mid x)$}
    \KwIn{family $\mathcal{M}(x, x')$ of inference strategies targeting $q(s \mid x'; x)$}
    \KwIn{initial position $x$ and estimate $\hat{Z}_{x}$ of $\tilde{\pi}(x)$}
    \KwOut{next position $x'$ and estimate $\hat{Z}_{x'}$ of $\tilde{\pi}(x')$}
    \nl $(s, x') \sim q(s, x'; x)$\;
    \nl $w_{x'} \gets {\texttt{HME}}(q(\cdot, x'; x), s, \mathcal{M}(x, x'))$\;
    \nl $(\_, w_x) \gets \texttt{IMPORTANCE}(q(\cdot \mid x; x'), \mathcal{M}(x', x))$\;
    \nl $(\_, \hat{Z}_{x'}) \gets \texttt{IMPORTANCE}(\pi(\cdot \mid x'), \mathcal{S}(x'))$\;
    \nl $u \sim \text{Uniform}(0, 1)$\;
    \nl \If{$u < \text{min}(1, \frac{\hat{Z}_{x'}}{\hat{Z}_x}w_{x'}w_x)$}{
        \nl \Return{$(x', \hat{Z}_{x'})$}\;
    }
    \nl \Else{
        \nl \Return{$(x, \hat{Z}_x)$}\;
    }
    }
    \end{algorithm}
\end{wrapfigure}

    
    
    
When $q$'s marginal density is known exactly, the above algorithm recovers variants of Particle-Marginal MH~\citep{andrieu2010particle}, except instead of using SMC to marginalize $r$, any RAVI algorithm can be applied. When $q$'s marginal density is unavailable, however, the algorithm instead becomes a pseudo-marginal \textit{ratio} algorithm~\citep{andrieu2018utility}, because not just $p$ but also $q$ is estimated unbiasedly. In general, it is \textit{not} valid to use arbitrary unbiased estimates of $p$ \textit{and} $q$, or even of $\alpha = \frac{p(x')q(x; x')}{p(x)q(x'; x)}$, within an MH algorithm. However, the added structure of the RAVI strategy ensures that the above procedure is sound.

To see why our MCMC kernel is stationary, we consider an extended target distribution. First, some notation. For an inference strategy $\mathcal{S}$ targeting $\pi(x)$, write $v_\mathcal{S}$ for the complete set of auxiliary variables in the strategy: if $\mathcal{S}.q$ has a tractable marginal density, then $v_\mathcal{S} = \emptyset$, and otherwise, if $\mathcal{S}.q(x) = \int \mathcal{S}.q(r, x) \text{d}r$, then $v_\mathcal{S}$ is defined recursively as $\{r\} \cup v_{\mathcal{S}.\mathcal{M}}$. Calling $\texttt{IMPORTANCE}$ on $\mathcal{S}$ yields a joint distribution over these auxiliary variables and $x$, which we denote as $p^{\mathcal{S}}_{\texttt{IMP}}(v_\mathcal{S}, x)$. Calling $\texttt{HME}$ on $\mathcal{S}$ and a particular sample $x$ yields a distribution over just $v_\mathcal{S}$, which we denote $p^\mathcal{S}_{\texttt{HME}}(v_\mathcal{S}; x)$.
When $x \sim \pi$ and $v_\mathcal{S} \sim p^\mathcal{S}_\texttt{HME}(v_\mathcal{S}; x)$, the ratio $\frac{p^\mathcal{S}_\texttt{IMP}(v_\mathcal{S}, x)}{\tilde{\pi}(x)p^\mathcal{S}_\texttt{HME}(v_\mathcal{S}; x)}$ is the weight $\check{Z}(\tilde{\pi}, \mathcal{S})^{-1}$ returned by \texttt{HME}, and similarly, when $(v_\mathcal{S}, x) \sim p^\mathcal{S}_\texttt{IMP}$, the ratio $\frac{\tilde{\pi}(x)p^\mathcal{S}_\texttt{HME}(v_\mathcal{S}; x)}{p^\mathcal{S}_\texttt{IMP}(v_\mathcal{S}, x)}$ is the weight $\hat{Z}(\tilde{\pi}, \mathcal{S})$ returned by $\texttt{IMPORTANCE}$.

Using this notation, we can extend the target distribution $\tilde{\pi}(x)$ to one over $(x, s, x', s', v_{\mathcal{S}(x)}, v_{\mathcal{M}(x, x')}, v_{\mathcal{M}(x', x)})$ that admits $\tilde{\pi}(x)$ as a marginal:
$$\tilde{\pi}(r, x, s, x', s', v_{\mathcal{S}(x)}, v_{\mathcal{M}(x, x')}, v_{\mathcal{M}(x', x)}) = \tilde{\pi}(r, x) \cdot p^{\mathcal{S}(x)}_\texttt{HME}(v_{\mathcal{S}(x)}; r) \cdot q(s, x'; x) \cdot p^{\mathcal{M}(x, x')}_\texttt{HME}(v_{\mathcal{M}(x, x')}; s) \cdot p^{\mathcal{M}(x', x)}_\texttt{IMP}(v_{\mathcal{M}(x', x)}, s')$$

Our algorithm can be understood as sequencing two stationary kernels for this extended target. The first (implemented by lines 1-3) is a blocked Gibbs update on the variables $(s, x', s', v_{\mathcal{M}(x, x')}, v_{\mathcal{M}(x', x)})$, conditioned on everything else. Lines 1-3 sample exactly from the conditional distribution of these variables. The second is a Metropolis-Hastings proposal that simultaneously: (i) swaps $x$ with $x'$ (the `main' proposed update), (ii) swaps $(s, v_{\mathcal{M}(x, x')})$ with $(s', v_{\mathcal{M}(x', x)})$, and (iii) proposes an update to $r$ and to $v_{\mathcal{S}(x)}$ from $p_\texttt{IMP}^{\mathcal{S}(x')}$. The usual Metropolis-Hastings acceptance probability for this kernel, computed on the extended state space, is precisely the formula in Line 6.

One consequence of this justification is that the \textit{same} family $\mathcal{S}$ of inference strategies for $\pi$ must be used at each iteration. The family $\mathcal{M}$ can be freely switched out (as can $q$), however, to develop a cycle of kernels that use different proposal distributions.

\section{Further Examples}
\label{sec:appendix-examples}

This appendix lists examples of popular Monte Carlo and variational inference algorithms, and explains how they can be viewed as inference strategies. In addition, some of these algorithms can be viewed as \textit{inference strategy combinators}, because they feature user-chosen proposal distributions or variational families that can themselves be instantiated with inference strategies.\footnote{This `combinator' viewpoint evokes earlier work by~\citep{scibior2018denotational} and \citep{stites2021learning}. For example, \citep{stites2021learning} introduce combinators for creating properly weighted samplers compositionally, with parameters that can be optimized using standard or nested variational objectives. Some of their combinators have equivalents in this section, e.g. their \texttt{propose} combinator is similar to the construction we present for Nested Importance Sampling in Section~\ref{sec:nsmc-example}. However: (1) the fundamental compositional operation in RAVI, of combining a posterior approximation with a meta-posterior approximation, cannot be achieved using their combinators; (2) as such, some of the algorithms that RAVI covers cannot be constructed using their combinators; and (3) their combinators produce properly weighted samplers, which contain `less information' than inference strategies: an inference strategy can be used, e.g., as a proposal distribution in Metropolis-Hastings, whereas properly weighted samplers cannot in general be used this way.}



\subsection{$N$-particle Importance Sampling}
\label{sec:sir-example}

\begin{wrapfigure}{L}{0.5\textwidth}
\vspace{-6mm}
\begin{algorithm}[H]
    \label{infstrat:sir}
    \SetAlgoLined\DontPrintSemicolon
    \footnotesize{
    \textbf{RAVI Inference Strategy:} $N$-particle Importance Sampling\;
    \SetKwFunction{sir}{sir($\tilde{\pi}, q, N$).q}\SetKwFunction{sirm}{sir($\tilde{\pi}, q, N$).M($x$).q}
    \SetKwProg{infalg}{Posterior Approx.}{}{}
    \SetKwInOut{Infers}{Target of inference}
    \SetKwInOut{Aux}{Auxiliary variables}
    \infalg{\sir{}}{
    \Infers{latent variable $x$}
    \Aux{particles $x_{1:N}$, chosen particle index $j$}
    \nl \For{$i \in 1, \dots, N$}{
        \nl $x_i \sim q$\;
        \nl $w_i \gets \frac{\tilde{\pi}(x_i)}{q(x_i)}$\;
    }
    \nl $j \sim \text{Discrete}(w_{1:N})$\;
    \nl \Return{$x_j$}\;}{}
    \setcounter{AlgoLine}{0}
    \SetKwProg{metaalg}{Meta-Posterior Approx.}{}{}
    \metaalg{\sirm{}}{
    \Infers{particles $x_{1:N}$, chosen particle index $j$}
    \Aux{None}
    \nl $j \sim \text{Uniform}(1, N)$\;
    \nl $x_j \gets x$\;
    \nl \For{$i \in 1,\dots,j-1,j+1,\dots,N$}{
        \nl $x_i \sim q$\;
    }
    \nl \Return{$(x_{1:N}, j)$}}
    }
    \vspace{-10mm}
\end{algorithm} 
\end{wrapfigure}

Consider the $N$-particle importance sampling estimator $$\hat{Z} = \frac{1}{N} \sum_{i=1}^N \frac{\tilde{\pi}(x_i)}{q(x_i)},\text{ for }x_i \sim q.$$ The same estimator can be recovered as a \textit{one-particle} \texttt{IMPORTANCE} estimate, by applying Alg.~\hyperref[alg:alg1]{1} to the~\hyperref[infstrat:sir]{\texttt{sir}} inference strategy.

The proposal $\mathcal{S}.q$ generates $N$ particles $x_{1:N}$, and selects an index $j$ from a discrete distribution on $1, \dots, N$, with weights proportional to $w_i = \tilde{\pi}(x_i) / q(x_i)$. The meta-proposal is responsible for inferring $j$ and the complete set of particles $x_{1:M}$, given the chosen particle $x$. It uses the conditional SIR algorithm~\citep{andrieu2010particle} to do so, proposing $j$ uniformly in $\{1, \dots, N\}$, and generating values for the un-chosen particles $x_{-j}$ from $q$.

\begin{wrapfigure}{L}{0.5\textwidth}
    \vspace{-6mm}
    \begin{algorithm}[H]
        \label{infstrat:ravisir}
        \SetAlgoLined\DontPrintSemicolon
        \footnotesize{
        \textbf{RAVI Inference Strategy:} $N$-particle IS with RAVI strategy $\mathcal{S}$\;
        \SetKwFunction{ravisir}{ravi-sir($\tilde{\pi}, \mathcal{S}, N$).q}\SetKwFunction{ravisirm}{ravi-sir($\tilde{\pi}, \mathcal{S}, N$).M($x$).q}
        \SetKwProg{infalg}{Posterior Approx.}{}{}
        \SetKwInOut{Infers}{Target of inference}
        \SetKwInOut{Aux}{Auxiliary variables}
        \infalg{\ravisir{}}{
        \Infers{latent variable $x$}
        \Aux{particles $x_{1:N}$, aux. proposal variables $v_\mathcal{S}^{1:N}$, chosen particle index $j$}
        \nl \For{$i \in 1, \dots, N$}{
            \nl $x_i, w_i \sim \texttt{IMPORTANCE}(\tilde{\pi}, \mathcal{S})$ w. aux. vars $v_\mathcal{S}^i$\;
        }
        \nl $j \sim \text{Discrete}(w_{1:N})$\;
        \nl \Return{$x_j$}\;}{}
        \setcounter{AlgoLine}{0}
        \SetKwProg{metaalg}{Meta-Posterior Approx.}{}{}
        \metaalg{\ravisirm{}}{
        \Infers{particles $x_{1:N}$, aux. proposal variables $v_\mathcal{S}^{1:N}$, chosen particle index $j$}
        \Aux{None}
        \nl $j \sim \text{Uniform}(1, N)$\;
        \nl $x_j \gets x$\;
        \nl $\_ \sim \texttt{HME}(\tilde{\pi}, x_j, \mathcal{S})$ w. aux. vars $v_\mathcal{S}^j$\;
        \nl \For{$i \in 1,\dots,j-1,j+1,\dots,N$}{
            \nl $\_, x_i \sim q$ w. aux. vars $v_{\mathcal{S}}^i$\;
        }
        \nl \Return{$(v_\mathcal{S}^{1:N}, x_{1:N}, j)$}}
        }
        \vspace{-5mm}
    \end{algorithm} 
\end{wrapfigure}

This is a suboptimal choice of $\mathcal{S}.\mathcal{M}(x).q$; lower-variance estimates $\hat{Z}$ can be obtained by improving meta-inference, either by incorporating problem-specific domain knowledge or via learning. However, in many cases, improved meta-inference may not be worth the computation required; it remains to be seen whether techniques such as amortized learning can be applied to deliver accuracy gains at low computational cost.

\textbf{Instantiating the proposal $q$ as its own inference strategy.} 
The above assumes that $q$ has a tractable marginal density. When it doesn't, the inner importance sampling loop can use a RAVI inference strategy $\mathcal{S}$ instead of a tractable proposal $q$. This modification is presented in the higher-order inference strategy~\hyperref[infstrat:ravisir]{\texttt{ravi-sir}}. 
One way to think about this construction is as a way to improve any existing inference strategy $\mathcal{S}$ by `adding replicates.' The resulting estimator of $Z$ is the mean of $N$ independent $\hat{Z}$ estimates from the original inference strategy.

\subsection{Importance-Weighted Autoencoders}
\label{sec:iwae-example}

The importance-weighted auto-encoder arises by considering the same inference strategy as in Section~\ref{sec:sir-example}, but as a variational inference procedure (Alg. 3) rather than a Monte Carlo procedure. 

Because $\texttt{sir}(\tilde{\pi}, q, N).q$ of this inference strategy corresponds to $N$-particle sampling importance-resampling (SIR), it has been argued that IWAE is in fact `vanilla' variational inference, but with a variational family that uses SIR to more closely approximate the posterior~\cite{bachman2015training}. However, \citep{cremer2017reinterpreting} show that deriving the ELBO for that variational family gives rise to a different objective, and that IWAE gives a looser lower bound on $\log Z$ than this idealized (but generally intractable) objective. 

In the RAVI framework, these two objectives arise from different inference strategies, which share the same $\mathcal{S}.q$ (SIR in both cases), but use different meta-inference $\mathcal{S}.\mathcal{M}$. IWAE uses the simple conditional SIR meta-inference introduced in Section~\ref{sec:sir-example}, whereas \citep{cremer2017reinterpreting}'s idealized objective can be derived by using the optimal choice of $\mathcal{S}.\mathcal{M}(x).q(j, x_{1:N})$\textemdash the exact posterior of the SIR procedure. The looser bound obtained by IWAE can be seen as a result of its $\mathcal{S}.\mathcal{M}$ performing poorer meta-inference: inference about the auxiliary variables of the SIR inference algorithm used in $\mathcal{S}.q$.

\subsection{$N$-particle Sequential Monte Carlo}
\label{sec:smc-example}

\begin{wrapfigure}{L}{0.6\textwidth}
    \vspace{-2mm}
    \begin{algorithm}[H]
        \label{infstrat:smc}
        \SetAlgoLined\DontPrintSemicolon
        \footnotesize{
        \textbf{RAVI Inference Strategy:} $N$-particle SMC w. RAVI strategies\;
        \SetKwFunction{ravismc}{smc($\tilde{\pi}_{1:T}, \mathcal{S}, K_{2:T}, L_{2:T}, N$).q}\SetKwFunction{ravismcm}{smc($\tilde{\pi}_{1:T}, \mathcal{S}, K_{2:T}, L_{2:T}, N$).M($x$).q}
        \SetKwProg{infalg}{Posterior Approx.}{}{}
        \SetKwInOut{Infers}{Target of inference}
        \SetKwInOut{Aux}{Auxiliary variables}
        \infalg{\ravismc{}}{
        \Infers{latent variable $x$ targeting $\tilde{\pi}_T$}
        \Aux{particles $x^{1:T}_{1:N}$, aux. proposal variables $v_\mathcal{S}^{1:N}$, aux. $K$ vars $v_{K_{2:T}}^{1:N}$, aux. $L$ vars $v_{L_{2:T}}^{1:N}$, ancestor variables $a^{1:T-1}_{1:N}$, final chosen particle index $j$}
        \nl \For{$i \in 1, \dots, N$}{
            \nl $x^1_i, w^1_i \sim \texttt{IMPORTANCE}(\tilde{\pi}_1, \mathcal{S})$ w. aux. vars $v_\mathcal{S}^i$\;
        }
        \nl \For{$t \in 2, \dots, T$}{
            \nl \For{$i \in 1, \dots, N$}{
                \nl $a_i^{t-1} \sim \text{Discrete}(w^{t-1}_{1:N})$\;
                \nl $x_i^t, \hat{w} \sim \texttt{IMPORTANCE}(\tilde{\pi}_t, K_t(x^{t-1}_{a^{t-1}_i}))$ w. aux. vars $v_{K_t}^i$\;
                \nl $\check{w} \sim \texttt{HME}(\tilde{\pi}_{t-1}, x^{t-1}_{a^{t-1}_i}, L_t(x_i^t))$ w. aux. vars $v_{L_t}^i$\;
                \nl $w_i^{t} \gets \hat{w} \cdot \check{w}$\;
            }
        }
        \nl $j \sim \text{Discrete}(w^T_{1:N})$\;
        \nl \Return{$x^T_j$}\;}{}
        \setcounter{AlgoLine}{0}
        \SetKwProg{metaalg}{Meta-Posterior Approx.}{}{}
        \metaalg{\ravismcm{}}{
        \Infers{particles $x^{1:T}_{1:N}$, aux. proposal variables $v_\mathcal{S}^{1:N}$, aux. $K$ vars $v_{K_{2:T}}^{1:N}$, aux. $L$ vars $v_{L_{2:T}}^{1:N}$, ancestor variables $a^{1:T-1}_{1:N}$, final chosen particle index $j$}
        \Aux{None}
        \nl $j \sim \text{Uniform}(1, N)$\;
        \nl $x_j^T, b_T \gets x, j$\;
        \nl \For{$t \in T, \dots, 2$}{
            \nl $a_{b_t}^{t-1} \sim \text{Uniform}(1, N)$\;
            \nl $b_{t-1} \gets a_{b_t}^{t-1}$\;
            \nl $x_{b_{t-1}}^{t-1}, \check{w} \sim \texttt{IMPORTANCE}(\tilde{\pi}_{t-1}, L_t(x_{b_t}^t))$ w. aux. vars $v_{L_t}^{b_t}$\;
            \nl $\hat{w} \sim \texttt{HME}(\tilde{\pi}_{t}, x_{b_t}^t, K_t(x_{b_{t-1}}^{t-1}))$ w. aux. vars $v_{K_t}^{b_t}$\;
            \nl $w_{b_t}^t \gets (\hat{w} \cdot \check{w})^{-1}$\;
        }
        $w^1_{b_1} \sim \texttt{HME}(\tilde{\pi}_1, x_{b_1}^1, \mathcal{S})$ w. aux. vars $v_\mathcal{S}^{b_1}$\;
        \nl \For{$i \in 1, \dots, b_1-1, b_1+1, \dots, N$}{
            \nl $x_{i}^1, w_i^1 \sim \texttt{IMPORTANCE}(\tilde{\pi}_1, \mathcal{S})$ w. aux. vars $v_\mathcal{S}^i$\;
        }
        \nl \For{$t \in 2, \dots, T$}{
            \nl \For{$i \in 1, \dots, b_t-1, b_t+1, \dots, N$}{
                \nl $a_i^{t-1} \sim \text{Discrete}(w^{t-1}_{1:N})$\;
                \nl $x_i^t, \hat{w} \sim \texttt{IMPORTANCE}(\tilde{\pi}_t, K_t(x^{t-1}_{a^{t-1}_i}))$ w. aux. vars $v_{K_t}^i$\;
                \nl $\check{w} \sim \texttt{HME}(\tilde{\pi}_{t-1}, x^{t-1}_{a^{t-1}_i}, L_t(x_i^t))$ w. aux. vars $v_{L_t}^i$\;
                \nl $w_i^{t} \gets \hat{w} \cdot \check{w}$\;
            }
        }
        \nl \Return{$(x_{1:N}^{1:T}, v_\mathcal{S}^{1:N}, v_{K_{2:T}}^{1:N}, v_{L_{2:T}}^{1:N}, a_{1:N}^{1:T-1},j)$}}
        }
        \vspace{-25mm}
    \end{algorithm} 
\end{wrapfigure}

The sequential Monte Carlo family of algorithms~\citep{chopin2020introduction, del2006sequential} evolve a population of \textit{weighted particles} to approximate a sequence of target distributions. SMC can be viewed as standard importance sampling, with an inference strategy in which $\mathcal{S}.q$ is the sampling distribution for SMC, and $\mathcal{S}.\mathcal{M}(x)$ is the conditional SMC algorithm~\citep{andrieu2010particle}.

Standard SMC is parameterized by:

\begin{enumerate}
    \item A sequence $\tilde{\pi}_{1:T}$ of intermediate target distributions, with $\tilde{\pi}_T = \tilde{\pi}$ the ultimate target;
    \item An initial proposal $q(x_1)$;
    \item A sequence $K_t(x_{t-1} \rightarrow x_t)$ of proposal kernels for $t=2, \dots, T$; and
    \item A sequence $L_t(x_{t} \rightarrow x_{t-1})$ of backward kernels for $t=2, \dots, T$.
\end{enumerate}

Here, we show a version of SMC (the inference strategy~\hyperref[infstrat:smc]{\texttt{smc}}) that behaves as a `higher-order inference strategy,' or `inference strategy combinator': it allows for an initial proposal, proposal kernels, and backward kernels that do not have tractable marginal densities. Our version is parameterized by:

\begin{enumerate}
    \item A sequence $\tilde{\pi}_{1:T}$ of intermediate target distributions, with $\tilde{\pi}_T = \tilde{\pi}$ the ultimate target;
    \item An initial proposal $\mathcal{S}$ (a RAVI strategy);
    \item A sequence of inference strategy families $K_t(x_{t-1})$ parameterized by $x_{t-1}$, for $t=2, \dots, T$, targeting $\tilde{\pi}_t$; and
    \item A sequence of inference strategy families $L_t(x_{t})$ of backward kernels, parameterized by $x_t$, for $t=2, \dots, T$.
\end{enumerate}

The posterior approximation $\mathcal{S}.q$ runs a version of SMC that uses $\texttt{HME}$ and $\texttt{IMPORTANCE}$ to compute weights. The meta-posterior approximation $\mathcal{S}.\mathcal{M}(x).q$ runs a similarly modified version of conditional SMC~\citep{andrieu2010particle}. When \texttt{IMPORTANCE} is run on the \texttt{smc} inference strategy, the final weight $\hat{Z}$ is the SMC marginal likelihood esitmate, the product of the averages of the weights from each time step.

It is possible to adapt this strategy to use adaptive resampling and rejuvenation. (Rejuvenation moves do not actually require modification: can be incorporated by including them as explicit $(K, L)$ pairs, where $L$ is the time-reversal of an MCMC kernel $K$.) However, we are not aware of a way to justify the adaptive choice of rejuvenation kernel.

\subsection{Variational Sequential Monte Carlo}
\label{sec:vsmc-example}

The Variational Sequential Monte Carlo~\citep{naesseth2018variational} objective corresponds to Alg. 3, with the same RAVI inference strategy as in Appendix~\ref{sec:smc-example}. However, the default gradient estimator from Alg. 3 will have high variance. \citet{naesseth2018variational} recommend using a biased estimator of the gradient, that uses reparameterization where possible and discards the score function terms arising from resampling steps.

\subsection{Annealed Importance Sampling}
\label{sec:ais-example}

\begin{wrapfigure}{L}{0.5\textwidth}
    \vspace{-6mm}
    \begin{algorithm}[H]
        \label{infstrat:ais}
        \SetAlgoLined\DontPrintSemicolon
        \footnotesize{
        \textbf{RAVI Inference Strategy:} Annealed Importance Sampling\;
        \SetKwFunction{ais}{ais($\tilde{\pi}_{1:T}, \mathcal{S}, K_{2:T}$).q}\SetKwFunction{aism}{ais($\tilde{\pi}_{1:T}, \mathcal{S}, K_{2:T}$).M($x$).q}
        \SetKwProg{infalg}{Posterior Approx.}{}{}
        \SetKwInOut{Infers}{Target of inference}
        \SetKwInOut{Aux}{Auxiliary variables}
        \infalg{\ais{}}{
        \Infers{latent variable $x$ targeting $\tilde{\pi}_T$}
        \Aux{$x^{1:T}$, aux. vars $v_\mathcal{S}$ of initial proposal}
        \nl $x_1, \_ \sim \texttt{IMPORTANCE}(\tilde{\pi}_1, \mathcal{S})$ w. aux. vars $v_\mathcal{S}$\;
        \nl \For{$t \in 2, \dots, T$}{
            \nl $x_t \sim K_t(x_{t-1} \rightarrow \cdot)$\;
        }
        \nl \Return{$x_T$}\;}{}
        \setcounter{AlgoLine}{0}
        \SetKwProg{metaalg}{Meta-Posterior Approx.}{}{}
        \metaalg{\aism{}}{
        \Infers{$x^{1:T}$, aux. vars $v_\mathcal{S}$ of initial proposal}
        \Aux{None}
        \nl $x_T \gets x$\;
        \nl \For{$t \in T, \dots, 2$}{
            \nl $x_{t-1} \sim \tilde{K}_t(x_t \rightarrow \cdot)$\tcp*{$\tilde{K}_t$ is time reversal of $K_t$}
        }
        \nl $\_ \sim \texttt{HME}(\tilde{\pi}_1, x_1, \mathcal{S})$ w. aux. vars $v_\mathcal{S}$\;
        \nl \Return{$(x_{1:T}, v_\mathcal{S})$}}
        }
        \vspace{-7mm}
    \end{algorithm} 
\end{wrapfigure}

In annealed importance sampling, the practitioner chooses a sequence of unnormalized target distributions $\tilde{\pi}_{1:T}$, where $\pi_T$ is the posterior distribution of interest. Typically $\pi_1$ is chosen to be a distribution that is easy to approximate with a proposal $q$, and each $\pi_i$ is slightly closer to the true target $\pi_T$ than the last. The user also chooses a sequence of kernels $K_t(x_{t-1} \rightarrow x_t)$, where $K_t$ is stationary for $\pi_{t-1}$. The algorithm begins by sampling an initial point $x_1 \sim q$, transforming it through the sequence of kernels to obtain $x_2, \dots, x_T$, and returning $x_T$ as the inferred value of $x$. The associated weight is $$\hat{Z} = \frac{\tilde{\pi}_1(x_1) \cdot \dots \cdot \tilde{\pi}_T(x_T)}{q(x_1) \cdot \tilde{\pi}_1(x_2) \cdot \dots \cdot \tilde{\pi}_{T-1}(x_T)}.$$

This procedure corresponds to running Alg.~\hyperref[alg:alg1]{1} on the~\hyperref[infstrat:ais]{\texttt{ais}} inference strategy. The inference process runs the kernels $K_t$ forward, whereas the meta-inference process runs their time reversals backward: $\tilde{K}_t(x_t \rightarrow x_{t-1}) \propto \pi_t(x_{t-1}) \cdot K_t(x_{t-1} \rightarrow x_t)$.

Note that if $K$ is a stationary kernel for $\pi_i$, so is $K^m$ for any natural number $m$. With sufficient computation (increasing $m$), we can ensure that the AIS top-level proposal $\texttt{ais}(\dots).q$ is arbitrarily close to the target posterior $\pi_T$. However, doing so will not necessarily lead to lower-variance weights: RAVI makes clear that it is also necessary to consider the quality of meta-inference. 

Consider the job of $\tilde{K}_T$, which in the context of the meta-posterior approximation $\texttt{ais}.\mathcal{M}(x)$ is supposed to infer $x_{T-1}$ from $x_{T}$. $\tilde{K}_T$  is the exact meta-posterior of $x_{T-1}$ given $x_T$ \textit{assuming that, in the forward direction, $x_{T-1}$ was distributed according to $\pi_{T-1}$}. However, in the forward direction, if each $K_t$ is run sufficiently many times to ensure mixing at each step, $x_{T-1}$ will in fact be distributed according to $\pi_{T-2}$. This gap\textemdash between the optimal meta-inference kernels and the actual $\tilde{K}$ kernels\textemdash is partly responsible for the variance of the AIS estimator, and can be mitigated by using a finer annealing schedule that brings successive target distributions closer together.  It could also be mitigated by learning a better reverse annealing chain.

\subsection{Nested Sequential Monte Carlo}
\label{sec:nsmc-example}

We first consider Nested Importance Sampling. As in RAVI, Nested Importance Sampling is concerned with importance sampling when the proposal distribution $q$  cannot be tractably evaluated. But RAVI and NIS take different approaches:

\begin{enumerate}
    \item RAVI assumes $q$ can be simulated, but that the (normalized) density cannot be evaluated. RAVI generates proposals exactly distributed according to the user's desired proposal $\mathcal{S}.q$, and generates approximations to the ideal importance weights.
    
    \item NIS does not assume $q$ can be simulated, but does assume that its unnormalized density $\tilde{q}$ is available. As such, proposals are not simulated from $q$, but rather from a Sampling/Importance-Resampling (SIR) approximation to $q$.
\end{enumerate}

The NIS procedure with an intractable proposal $q$ corresponds exactly to a special case of the RAVI algorithm, with the RAVI proposal $\mathcal{S}.q$ set \textit{not} to $q$ but rather to an SIR sampling distribution targeting $q$ using some tractable proposal $h$. Compare:

\begin{itemize}
    \item Ordinary SIR targeting $\tilde{\pi}$ with proposal $h$: recovered by running $\texttt{IMPORTANCE}(\tilde{\pi}, \texttt{sir}(\tilde{\pi}, h, N))$ (see Section~\ref{sec:sir-example} for \texttt{sir} inference strategy).
    \item Nested IS targeting $\tilde{\pi}$ with unnormalized proposal density $\tilde{q}$, approximated using SIR with $h$ as a proposal: recovered by running $\texttt{IMPORTANCE}(\tilde{\pi}, \texttt{sir}(\tilde{q}, h, N))$.
\end{itemize}

That is, under the RAVI perspective, the only difference between ordinary SIR using $h$ and nested IS is that the ideal proposal density $\tilde{q}$ (rather than the target density $\tilde{\pi}$) is used to make the resampling decision about the particles generated by $h$ (the index $j$ in the listing for \texttt{sir}).


More generally, \citet{naesseth2015nested} consider procedures other than SIR for approximating ${q}$, arguing that any properly weighted sampler for the intractable proposal $q$ will do. If we let $\mathcal{H}$ be a RAVI inference strategy representing the properly weighted sampler for the intractable proposal $q$ (with unnormalized density $\tilde{q}$), then the Nested IS procedure that uses this properly weighted proposal to perform inference in $\tilde{\pi}$ is $\texttt{IMPORTANCE}(\tilde{\pi}, \texttt{ravi-sir}(\tilde{q}, \mathcal{H}, 1))$ (see~\hyperref[infstrat:ravisir]{\texttt{ravi-sir}} in Section~\ref{sec:sir-example}).


Nested SMC is similar, performing Nested IS at each iteration of SMC. To recover this algorithm using RAVI, we use the~\hyperref[infstrat:smc]{\texttt{smc}} inference strategy, but for the proposals $K_t(x_{t-1})$ (which, as described in Section~\ref{sec:smc-example}, can be instantiated with inference strategies), we use~\hyperref[infstrat:ravisir]{\texttt{ravi-sir}} targeting the desired but intractable proposal.






\subsection{SMC$^2$}
\label{sec:smcsq-example}

\begin{wrapfigure}{L}{0.6\textwidth}
    \vspace{-6mm}
    \begin{algorithm}[H]
        \label{infstrat:smcsq}
        \SetAlgoLined\DontPrintSemicolon
        \footnotesize{
        \textbf{RAVI Inference Strategy:} SMC$^2$\;
        \SetKwFunction{smcsq}{smc$^2$($p, q_1, q, M, N$).q}\SetKwFunction{smcsqm}{smc$^2$($p, q_1, q, M, N$).M($\theta, x_{1:T}$).q}
        \SetKwProg{infalg}{Posterior Approx.}{}{}
        \SetKwInOut{Infers}{Target of inference}
        \SetKwInOut{Aux}{Auxiliary variables}
        \infalg{\smcsq{}}{
        \Infers{parameters $\theta$, sequence $x_{1:T}$}
        \Aux{inner SMC vars $v_\texttt{smc}^T$ of chosen SMC$^2$ particle, other SMC$^2$ vars $v$}
        \nl \tcp{the targets $\tilde{\pi}_t$ depend on $M$, $p$, $q_1$, and $q$}
        \nl $(\theta, x_{1:T}, v_\texttt{smc}^T), \_ \sim \texttt{IMPORTANCE}(\tilde{\pi}_T, \texttt{smc}(\tilde{\pi}_{1:T}, K_{2:T}^2, L_{2:T}^2, N))$ w. aux. vars $v$\;
        \nl \Return{$\theta, x_{1:T}$}\;}{}
        \setcounter{AlgoLine}{0}
        \SetKwProg{metaalg}{Meta-Posterior Approx.}{}{}
        \metaalg{\smcsqm{}}{
        \Infers{inner SMC vars $v_\texttt{smc}^T$ of chosen SMC$^2$ particle, other SMC$^2$ vars $v$}
        \Aux{None}
        \nl $\_ \sim \texttt{HME}(p_T^\theta, x_{1:T}, \texttt{smc}(p^\theta_{1:T}, q_1, K_{2:T}, L_{2:T}, M))$ w. aux. vars $v_\texttt{smc}^T$\;
        \nl $\_ \sim \texttt{HME}(\tilde{\pi}_T, (\theta, x_{1:T}, v_\texttt{smc}^T), \texttt{smc}(\tilde{\pi}_{1:T}, K_{2:T}^2, L_{2:T}^2, N))$ w. aux. vars $v$\;
        \nl \Return{$(v_\texttt{smc}^T, v)$}}
        }
        \vspace{-7mm}
    \end{algorithm} 
\end{wrapfigure}

Suppose we are working with a state-space model $p(\theta) \prod_{i=1}^T p(x_i \mid x_{1:i}, \theta) p(y_i \mid x_i, \theta)$. For a fixed $\theta$, an SMC algorithm could be used to target the successive posteriors $p_t^\theta(x_{1:t}) = p(x_{1:t} \mid y_{1:t}, \theta)$, with proposal kernels $K_t(x^{t-1}_{1:t-1} \rightarrow x^t_{1:t}) = \delta_{x^{t-1}_{1:t-1}}(x^t_{1:t-1})q(x^t_t; x^t_{1:t-1}, y_{1:t}, \theta)$ (for some choice of $q$) and deterministic backward kernels $L_t(x^t_{1:t} \rightarrow x^{t-1}_{1:t-1}) = \delta_{x^t_{1:t-1}}(x^{t-1}_{1:t-1})$. The RAVI strategy implementing that SMC algorithm is $\texttt{smc}(p_{1:T}, q_1, K_{2:T}, L_{2:T}, N)$, where $q_1(x_1; \theta)$ is a proposal for an initial $x_1$ and $N$ is the number of particles. 

If we also wish to infer $\theta$, we can instead use the SMC$^2$ algorithm~\citep{chopin2013smc2}. We define extended targets
$$\pi_t(\theta, x_{1:t}, v_\texttt{smc}^t) = p(\theta \mid y_{1:t}) p(x_{1:t} \mid y_{1:t}, \theta) p_{\texttt{HME}}^{\texttt{smc}(p_{1:t}^\theta, q_1, K_{2:t}, L_{2:t}, N)}(v_\texttt{smc}^t; x_{1:t}),$$
which are defined over not only $\theta$ and $x_{1:t}$ but also all the auxiliary variables $v_\texttt{smc}^t$ used during steps 1 through $t$ of SMC. The variables $v_\texttt{smc}$ and the $p_\texttt{HME}$ distribution over them are as defined in Appendix~\ref{sec:ravi-mcmc}. We write $\tilde{\pi}_t$ for the unnormalized versions of these targets, with normalizing constant $p(y_{1:t})$.

The SMC$^2$ algorithm targets this sequence of extended posteriors. We write $K_t^2$ for the forward kernels used  by this outer SMC algorithm. The kernel $K_t^2$ extends the SMC state variables $v_\texttt{SMC}^{t-1}$ to new state variables $v_\texttt{SMC}^{t}$ by running the particle filter forward one step, resampling the chosen trajectory index $j$ based on the new weights for time step $t$, and updating $x_{1:t}$ to match the $j^\text{th}$ trajectory. The corresponding backward kernel $L_t^2$ deletes the $t^\text{th}$ step of the particle deterministically, then reproposes $j$ based on the step $t-1$ weights, setting $x_{1:t-1}$ to match the $j^\text{th}$ trajectory.

The SMC$^2$ algorithm corresponds to the RAVI strategy~\hyperref[infstrat:smcsq]{\texttt{smc$^2$}}. Running the other SMC yields an approximate sample from $\tilde{\pi}_T$, which includes auxiliary variables $v_\texttt{smc}^T$. Meta-inference runs two rounds of conditional SMC: first, to recover the inner layer of SMC's variables $v_\texttt{smc}^T$ for the chosen outer-layer particle, and second, to recover the outer layer of SMC's auxiliary variables $v$.
As discussed by~\citet{chopin2013smc2}, particle MCMC rejuvenation moves can also be included; to justify using RAVI, we would insert these kernels as additional proposals within the sequence $K_{2:T}^2$. 

\subsection{Amortized Rejection Sampling}
\label{sec:amrej-example}

Consider a generative model $p(K, x_{1:K+1}, y)$ where the latent variables $x_{1:K+1}$ to be marginalized or inferred represent the trace of a rejection sampling loop, with sampling distribution $h(x)$ and predicate $\mathcal{A}(x)$ determining acceptance:
$$p(K, x_{1:K+1}, y) = \prod_{i=1}^{K} \left[h(x_i)(1-\mathcal{A}(x_i))\right] h(x_{K+1})\mathcal{A}(x_{K+1}) p(y \mid x_{K+1})$$

\begin{wrapfigure}{L}{0.7\textwidth}
    \vspace{-6mm}
    \begin{algorithm}[H]
        \label{infstrat:amrej}
        \SetAlgoLined\DontPrintSemicolon
        \footnotesize{
        \textbf{RAVI Inference Strategy:} Amortized Rejection Sampling\;
        \SetKwFunction{amrej}{amrej($h, q, \mathcal{A}, N, M$).q}\SetKwFunction{amrejm}{amrej($h, q, \mathcal{A}, N, M$).M($K, x_{1:K}$).q}
        \SetKwFunction{amrejmm}{amrej($h, q, \mathcal{A}, N, M$).M($K, x_{1:K}$).M($K', x'_{1:K'}, (K''_i, x''^i_{1:K''_i})_{i=1:M}, j$).q}
        \SetKwFunction{amrejmmm}{amrej($h, q, \mathcal{A}, N, M$).M($K, x_{1:K}$).M($K', x'_{1:K'}, (K''_i, x''^i_{1:K''_i})_{i=1:M}, j$).M($z_{K'+1}$).q}
        \SetKwProg{infalg}{Posterior Approx.}{}{}
        \SetKwInOut{Infers}{Target of inference}
        \SetKwInOut{Aux}{Auxiliary variables}
        \infalg{\amrej{}}{
        \Infers{number $K$ of rejected samples, rejected samples $x_{1:K}$, accepted sample $x_{K+1}$}
        \Aux{rejection loops $(K', x'_{1:K'})$ and $(K''_i, x''^i_{1:K''_i})_{i=1:M}$, index $j$}
        \nl $K' \gets 0$\;
        \nl $x'_{1} \sim q$\;
        \nl \While{$\mathcal{A}(x'_{K'+1}) \neq 1$}{
            \nl $K' \gets K' + 1$\;
            \nl $x'_{K'+1} \sim q$
        }
        \nl $x_{K+1} \gets x'_{K'+1}$\;
        \nl \For{$i \in 1, \dots, M$}{
            \nl $K''_i \gets 0$\;
            \nl $x''^i_{1} \sim h$\;
            \nl \While{$\mathcal{A}(x''^i_{K''_i+1}) \neq 1$}{
                \nl $K''_i \gets K''_i + 1$\;
                \nl $x''^i_{K''_i+1} \sim h$\;
            }
        }
        \nl $j \sim \text{Discrete}(K''_{1:M})$\;
        \nl $K \sim \text{Uniform}(0, K''_j)$\;
        \nl $x_{1:K} \gets x''^j_{1:K}$\;
        \nl \Return{$(K, x_{1:K}, x_{K+1})$}\;}{}
        \setcounter{AlgoLine}{0}
        \SetKwProg{metaalg}{Meta-Posterior Approx.}{}{}
        \metaalg{\amrejm{}}{
        \Infers{rejection loops $(K', x'_{1:K'})$ and $(K''_i, x''^i_{1:K''_i})_{i=1:M}$, index $j$}
        \Aux{superfluous accepted sample $z_{K'+1}$}
        \nl $j \sim \text{Uniform}(1, M)$\;
        \nl \For{$i \in 1, \dots, j-1, j+1, \dots, M$}{
            \nl $K''_i \gets 0$\;
            \nl $x''^i_{1} \sim h$\;
            \nl \While{$\mathcal{A}(x''^i_{K''_i+1}) \neq 1$}{
                \nl $K''_i \gets K''_i + 1$\;
                \nl $x''^i_{K''_i+1} \sim h$\;
            }
        }
        \nl $K''_j \gets K$\;
        \nl $x''^j_{K+1} \sim h$\;
        \nl \While{$\mathcal{A}(x''^j_{K+K''_j+1}) \neq 1$}{
            \nl $K''_j \gets K''_j + 1$\;
            \nl $x''^j_{K+K''_j+1} \sim h$\;
        }
        \nl $K' \gets 0$\;
        \nl $z_{1} \sim q$\;
        \nl \While{$\mathcal{A}(z_{K'+1}) \neq 1$}{
            \nl $x'_{K'+1} \gets z_{K'+1}$\;
            \nl $K' \gets K' + 1$\;
            \nl $z_{K'+1} \sim q$
        }
        \nl \Return{$(K', x'_{1:K'}, (K''_i, x''^i_{1:K''_i})_{i=1:M}, j)$}}
        \setcounter{AlgoLine}{0}
        \SetKwProg{metaalg}{Meta-Meta-Posterior Approx.}{}{}
        \metaalg{\amrejmm{}}{
        \Infers{superfluous accepted sample $z_{K'+1}$}
        \Aux{index $l$, unchosen particles $z_{-l}$}
        \nl \For{$i \in 1, \dots, N$}{
            \nl $z_i \sim q$\;
        }
        \nl $l \sim \text{Uniform}(\{i \mid \mathcal{A}(z_i)\})$\;
        \nl \Return{$z_l$}}
        \setcounter{AlgoLine}{0}
        \SetKwProg{metaalg}{Meta-Meta-Meta-Posterior Approx.}{}{}
        \metaalg{\amrejmmm{}}{
        \Infers{index $l$, unchosen particles $z_{-l}$}
        \Aux{None}
        \nl $l \sim \text{Uniform}(1, N)$\;
        \nl \For{$i \in 1, \dots, l-1,l+1,\dots, N$}{
            \nl $z_i \sim q$\;
        }
        \nl \Return{$(z_1, \dots, z_{l-1}, z_{l+1}, \dots, z_N)$}}
        }
        \vspace{-15mm}
    \end{algorithm} 
\end{wrapfigure}

Here, the $x_i$ are drawn independently from a distribution $h$, until some predicate $\mathcal{A}$ holds of the most recent particle, at which point the loop stops. The observation $y$ depends on the final sample $x_{K+1}$, but not the earlier, rejected samples $x_{1:K}$ or the number of rejected samples $K$.
\citet{naderiparizi2019amortized} proposed a technique called \textit{Amortized Rejection Sampling} for performing inference in this model. The technique corresponds to the rather involved RAVI strategy \texttt{amrej}, which has parameters $N$ and $M$ that can be used to trade accuracy for computational cost. 

The idea behind the top-level, intractable posterior approximation $\texttt{amrej}(h, q, \mathcal{A}, N, M).q$ is to:

\begin{itemize}
    \item use the observation $y$ to intelligently guess the \textit{accepted} particle $x_{K+1}$, using a learned proposal $q$. (For example, $q$ may be parameterized by a neural network that accepts $y$ as input.) To satisfy the constraint that $x_{K+1}$ satisfies $\mathcal{A}$, however, it is necessary to run $q$ within a rejection sampling loop, generating auxiliary variables $x'_{1:K'}$, where $K'$ is the number of rejected $q$-samples. (We could try directly using $x'_{1:K'}$ as our proposal for $x_{1:K}$, the rejected samples from the model. But $q$'s goal is to propose $x_{K+1}$ in a \textit{data-driven} way, influenced by the observation $y$, and the rejected samples $x_{1:K}$ from the model have no connection to the data\textemdash so, using samples from $q$ as proposals for the rejected model samples would result in a poor approximation.)

    \item use rejection sampling from the prior $h$ to infer the \textit{rejected} samples $x_{1:K}$. We run $M$ independent rejection sampling loops, randomly choose one with probability proportional to its length, and then randomly choose a \textit{prefix} of the chosen loop as our proposal for $x_{1:K}$. 
\end{itemize}

The meta-posterior approximation must solve two new challenges: recovering the rejected $q$ samples $x'_{1:K'}$ from the posterior approximation, and recovering the many unused rejection loops (and the suffix of the chosen rejection loop) from the second step of the posterior approximation (the $x''$ variables). The latter of these tasks is simple enough: we can generate $M-1$ rejection loops from scratch for the un-chosen loops, and a further rejection loop from scratch to use as the suffix of the chosen loop. The first task is more complex: we run a new rejection loop using $q$ as a proposal, and discard the final accepted sample. Meta-meta-inference must infer this discarded accepted sample, for which it uses SIR with $N$ particles. The final layer, the Meta-Meta-Meta-Posterior Approximation, uses conditional SIR.

The meta-meta-posterior is not absolutely continuous with respect to its approximation (it is possible that the approximation generates $N$ $z$-values that all fail to satisfy the predicate, in which case $z_l$ is not in the support of the meta-meta-posterior). As such, this is an example of a \textit{wide} inference strategy (Appendix~\ref{sec:even-odd}).

\subsection{Hamiltonian Variational Inference}
\label{sec:ham-example}

\begin{wrapfigure}{L}{0.5\textwidth}
    \vspace{-6mm}
    \begin{algorithm}[H]
        \label{infstrat:ham}
        \SetAlgoLined\DontPrintSemicolon
        \footnotesize{
        \textbf{RAVI Inference Strategy:} Hamiltonian Variational Inference\;
        \SetKwFunction{ham}{hamvi($q_0, q_v, r_v, \text{LF}$).q}\SetKwFunction{hamm}{hamvi($q_0, q_v, r_v, \text{LF}$).M($x$).q}
        \SetKwFunction{hammm}{hamvi($q_0, q_v, r_v, \text{LF}$).M($x$).M($x_0, v$).q}
        \SetKwProg{infalg}{Posterior Approx.}{}{}
        \SetKwInOut{Infers}{Target of inference}
        \SetKwInOut{Aux}{Auxiliary variables}
        \infalg{\ham{}}{
        \Infers{latent variable $x$}
        \Aux{initial position $x_0$, momentum $v$}
        \nl $x_0 \sim q_0$\;
        \nl $v \sim q_v$\;
        \nl $(x, v') \gets \text{LF}(x_0, v)$\;
        \nl \Return{$x$}\;}{}
        \setcounter{AlgoLine}{0}
        \SetKwProg{metaalg}{Meta-Posterior Approx.}{}{}
        \metaalg{\hamm{}}{
        \Infers{initial position $x_0$, momentum $v$}
        \Aux{negated final momentum $v'_{-}$}
        \nl $v'_{-} \sim r_v(\cdot; x)$\;
        \nl $(x_0, v_{-}) \gets \text{LF}(x, v'_{-})$\;
        \nl \Return{$(x_0, -v_{-})$}}
        \setcounter{AlgoLine}{0}
        \SetKwProg{metaalg}{Meta-Posterior Approx.}{}{}
        \metaalg{\hammm{}}{
        \Infers{negated final momentum $v'_{-}$}
        \Aux{None}
        \nl $(\_, v') \gets \text{LF}(x_0, v)$\;
        \nl \Return{$-v'$}}
        }
        \vspace{-12mm}
    \end{algorithm} 
\end{wrapfigure}

Hamiltonian Variational Inference~\citep{salimans2015markov} is a hybrid of Hamiltonian Monte Carlo and variational inference. It is a special case of Markov Chain Variational Inference (see Section~\ref{sec:overview} and Section~\ref{sec:examples} for detailed discussion, and~\hyperref[infstrat:mcvi]{\texttt{mcvi}} for the RAVI implementation). The algorithm specializes the Markov Chain Variational Inference procedure for use with a Hamiltonian Monte Carlo kernel. 

We present the specialized strategy as~\hyperref[infstrat:ham]{\texttt{hamvi}}. It accepts as input:

\begin{enumerate}
    \item a distribution $q_0$ from which to propose an initial point;
    \item a momentum distribution $q_v$ from which momenta $v$ are proposed at each iteration;
    \item a proposal distribution $r_v(\cdot; x)$ over momenta; and
    \item a leapfrog integrator $\texttt{LF}$ that runs Hamiltonian dynamics on an initial position and momentum (we think of both the number of leapfrog steps $L$ and the Hamiltonian $H$ being targeted as part of the \texttt{LF} object provided to \texttt{hamvi}).
\end{enumerate}

Given these inputs, the top-level posterior approximation runs an iteration of HMC from a randomly initialized location $x_0$. The meta-posterior approximation randomly proposes a (negated) \textit{final} momentum from the proposal $r_v$, and runs the leapfrog integrator to find a plausible initial location $x_0$. Finally, the (deterministic) meta-meta-posterior finds the initial momentum that could have taken $x_0$ to $x$.

\subsection{Antithetic Sampling}
\label{sec:antithetic-example}

Consider a target $\tilde{\pi}(x)$ and a proposal $q(x)$ that approximates $\pi$. Suppose $q$ is invariant under some bijective transformation $T$: $$\forall x, q(x) = q(T(x)).$$ For example, a univariate Gaussian proposal with mean $\mu$ is invariant under $T(x) = 2\mu - x$. Antithetic sampling generates a sample $x$ from $q$, but instead of using the estimator $\hat{Z} = \tilde{\pi}(x)/q(x)$, it uses $$\hat{Z} = \frac{\tilde{\pi}(x) + \tilde{\pi}(T(x))}{2q(x)}.$$

\begin{wrapfigure}{L}{0.6\textwidth}
    \vspace{-6mm}
    \begin{algorithm}[H]
        \label{infstrat:antithetic}
        \SetAlgoLined\DontPrintSemicolon
        \footnotesize{
        \textbf{RAVI Inference Strategy:} Antithetic Sampling\;
        \SetKwFunction{antithetic}{antithetic($\tilde{\pi}, q, T$).q}\SetKwFunction{antitheticm}{antithetic($\tilde{\pi}, q, T$).M($x$).q}
        \SetKwProg{infalg}{Posterior Approx.}{}{}
        \SetKwInOut{Infers}{Target of inference}
        \SetKwInOut{Aux}{Auxiliary variables}
        \infalg{\antithetic{}}{
        \Infers{latent variable $x$}
        \Aux{sampled $x_0$, choice $b$}
        \nl $x_0 \sim q$\;
        \nl $w_0 \gets \tilde{\pi}(x_0) / q(x_0)$\;
        \nl $w_1 \gets \tilde{\pi}(T(x_0)) / q(x_0)$\;
        \nl $b \sim \texttt{Bernoulli}(\frac{w_1}{w_0 + w_1})$\;
        \nl \Return{$bT(x_0) + (1-b)x_0$}\;}{}
        \setcounter{AlgoLine}{0}
        \SetKwProg{metaalg}{Meta-Posterior Approx.}{}{}
        \metaalg{\antitheticm{}}{
        \Infers{sampled $x_0$, choice $b$}
        \Aux{None}
        \nl $b \sim \text{Bernoulli}(0.5)$\;
        \nl $x_0 \gets bT(x) + (1-b)x$\;
        \nl \Return{$(x_0, b)$}}
        }
        \vspace{-3mm}
    \end{algorithm} 
\end{wrapfigure}

This can be justified as Algorithm 1 (\texttt{IMPORTANCE}) applied to the strategy~\hyperref[infstrat:antithetic]{\texttt{antithetic}}. The posterior approximation generates an initial sample $x_0 \sim q$, evaluates both $x_0$ and $T(x_0)$ as possible proposals, and selects one. The meta-posterior approximation must recover whether $x$ or its transformed version was the sampled one; it does so by flipping a fair coin, which is optimal when $T = T^{-1}$, i.e., when $T$ is an involution. In the general case a lower-variance estimator could be derived by setting $\mathcal{M}(x).q$ to the exact posterior of the proposal process. Antithetic sampling can also be generalized to the case where a finite family of bijective transformations $T_i$ are available.

Note that although the final expression for $\hat{Z}$ falls out of this inference strategy only when $q(x) = q(T(x))$ for all $x$, nothing in the inference strategy itself exploits this assumption, and the same inference strategy could be applied to $T$ without this property, to derive other estimators that\textemdash intuitively\textemdash simultaneously consider a proposal $x$ and a deterministic function of it $T(x)$ as possible locations.

\section{Absolute continuity}
\label{sec:even-odd}

When we defined inference strategies $\mathcal{S}$ targeting $\pi$, we required that $\mathcal{S}.q$ and $\pi$ be \textit{mutually} absolutely continuous, a stronger requirement than in importance sampling. We now consider relaxing this requirement, by requiring only \textit{one-sided} absolute continuity. We define two \textit{kinds} of inference strategy, depending on which direction of absolute continuity holds:

\begin{enumerate}
    \item An inference strategy $\mathcal{S}$ targeting $\pi$ is \textit{wide} if $\pi$ is absolutely continuous with respect to $\mathcal{S}.q$, and either $\mathcal{S}.q$ has a tractable marginal density or $\mathcal{S}.\mathcal{M}(x)$ is a narrow inference strategy targeting $\mathcal{S}.q(\cdot \mid x)$ for all $x$.
    
    \item An inference strategy $\mathcal{S}$ targeting $\pi$ is \textit{narrow} if $\mathcal{S}.q$ is absolutely continuous with respect to $\pi$, and either $\mathcal{S}.q$ has a tractable marginal density or $\mathcal{S}.\mathcal{M}(x)$ is a wide inference strategy targeting $\mathcal{S}.q(\cdot \mid x)$ for all $x$.
\end{enumerate}

Then an inference strategy as defined in the main paper is one that is both wide and narrow. 

Narrow inference strategies can serve as variational families within variational inference algorithms. Wide inference strategies can be used as importance sampling and SMC proposals, as well as variational families for \textit{amortized} variational inference.  Inference strategies used as MCMC proposals must be both wide and narrow.

\section{Other applications of RAVI inference strategies}
\label{sec:other-applications}

\subsection{Rejection sampling with RAVI}
As in any properly weighted sampler, if the weights produced by Alg. 1 can be bounded above by a constant $M$, a RAVI inference strategy can be used for exact inference via rejection sampling: a sample $(x, \hat{Z})$ is drawn using Alg. 1, and then accepted with probability $\frac{\hat{Z}}{M}$. The weight $\hat{Z}$ for an inference strategy can be viewed as a product of the normalizing constant $Z$ with normalized importance weights $w_\mathcal{S} = \frac{\pi(x)}{\mathcal{S}.q(x)}$, $w_{\mathcal{S}.\mathcal{M}(x)} = \frac{\mathcal{S}.q(r \mid x)}{\mathcal{S}.\mathcal{M}(x).q(r)}$, and so on. As such, if upper bounds $M_Z$ and $M_\mathcal{S}$, $M_{\mathcal{S}.\mathcal{M}(x)}$, etc. can be found for these quantities, the product of these bounds is a bound on $\hat{Z}$. Thus, as in properly weighted sampling and in variational inference with RAVI, it is possible to reason about the RAVI inference strategy compositionally, in terms of bounds at each layer of nesting.

\subsection{Estimating KL divergences between models with RAVI inference strategies equipped}

Suppose $p(y) = \int p(x, y) \text{d}x$ and $q(y) = \int q(z, y) \text{d}z$ are mutually absolutely continuous distributions over some space $\mathcal{Y}$. Suppose also that we have two families of inference strategies, $\mathcal{S}_p(y)$ and $\mathcal{S}_q(y)$, targeting $p(x\mid y)$ and $q(z \mid y)$ respectively. Then the AIDE algortihm~\citep{cusumano2017aide} can be adapted to give a stochastic upper bound on the symmetric KL divergence between $p(y)$ and $q(y)$.

First, we generate $(x, y_p) \sim p$, $(z, y_q) \sim q$, and run $\texttt{HME}$ on each pair to obtain weights $w^p_p$ and $w^q_q$ respectively. Then, we run $\texttt{IMPORTANCE}$ on $p$ with data $y_q$, and on $q$ with data $y_p$, to obtain weights $w^p_q$ and $w^q_p$ respectively. Finally, we sum the logs of the foru weights, to give an estimate $\hat{D}$ whose expectation is:

$$
\mathbb{E}[\hat{D}] =
\mathbb{E}_{y \sim p}[\UU(p, y, \mathcal{S}_p(y)) - \LL(q, y, \mathcal{S}_q(y))] + \mathbb{E}_{y \sim q}[\UU(q, y, \mathcal{S}_q(y)) - \LL_p(p, y, \mathcal{S}_p(y))] \geq KL(p || q) + KL(q || p).
$$

As the marginal likelihood bounds $\UU$ and $\LL$ become tighter, this expectation approaches the true symmetric KL between $p$ and $q$, i.e., $D = KL(p|| q) + KL(q || p)$. Theorem 4 allows us to characterize the tightness of these bounds, and thus of the stochastic upper bound $\hat{D}$ on the symmetric KL, in terms of KL divergences between successive layers of each inference strategy. Improving inference at any layer of the inference strategy tightens the bound $\hat{D}$, yielding less biased estimates of $D$.

\section{Reparameterization Trick Gradient Estimators}
\label{sec:reparam}
In this section, we present versions of Algorithms 3 and 4 that 
utilize reparameterization gradients, rather than score function 
gradients. Using these algorithms requires that an inference 
strategy be \textit{reparameterizable}.

\textbf{Definition:} A \textit{reparameterizable inference strategy $\mathcal{S}$ with arguments $\theta$} specifies:

\begin{itemize}
\item A reparameterizable posterior approximation $\mathcal{S}.q$, which is one of:
\begin{itemize} 
    \item a tractable proposal: a tuple $(\mathcal{S}.q(x; \theta), \mathcal{S}.q.g(\epsilon), \mathcal{S}.q.f(\epsilon, \theta)$, such that $q$ is the pushforward of $g$ by $f$; or
    \item an intractable proposal: a tuple $(\mathcal{S}.q(r, x; \theta), \mathcal{S}.q.g(\epsilon_r, \epsilon_x), \mathcal{S}.q.f_r(\epsilon_r, \theta), \mathcal{S}.q.f_x(\epsilon_x, \theta))$, such that $q$ is the pushforward of $g$ by $\lambda (\epsilon_r, \epsilon_x). (f_r(\epsilon_r, \theta), f_x(\epsilon_x, \theta))$.
\end{itemize}
\item If the latter, a reparameterizable meta-inference strategy $\mathcal{S}.\mathcal{M}$, with arguments $(x, \theta)$, that given argument $(x, \theta)$, targets $\mathcal{S}.q(r \mid x; \theta)$.
\end{itemize}

Now, reparameterized estimators can be derived by applying standard automatic differentiation to the following algorithm, which only samples from distributions that do not depend on parameters:

\begin{minipage}[t]{0.52\textwidth}
    {
    \removelatexerror
    \vspace{-10pt}
    \begin{algorithm}[H]
        \label{alg:alg6}
        \SetAlgoLined\DontPrintSemicolon
        \textbf{Algorithm 6:} RAVI ELBO estimator ($\texttt{ELBO}$)\;
        \KwIn{unnormalized model $\tilde{p}(x)$}
        \KwIn{inference strategy $\mathcal{S}$ with arguments}
        \KwIn{arguments $\theta$}
        \KwOut{unbiased estimates of $\mathcal{L}$ (differentiable w.r.t. $\theta$)}
        \nl \If{$\mathcal{S}.q$ has a tractable marginal density}{
            \nl $\epsilon_x \sim \mathcal{S}.q.g$\;
            \nl $x \gets \mathcal{S}.q.f(\epsilon_x, \theta)$\;
            \nl $\hat{U} \gets \log \mathcal{S}.q(x; \theta)$\;
        }
        \nl \ElseIf{$\mathcal{S}.q(x; \theta) = \int \mathcal{S}.q(r, x; \theta)\text{d}r$}{
            \nl $(\epsilon_r, \epsilon_x) \sim \mathcal{S}.q.g$\;
            \nl $(x, r) \gets (\mathcal{S}.q.f_x(\epsilon_x, \theta), \mathcal{S}.q.f_r(\epsilon_r, \theta))$\;
            \nl $\hat{U} \gets \texttt{EUBO}(\mathcal{S}.q(\cdot, x; \theta), r, \mathcal{S}.\mathcal{M}, (x, \theta))$\;
        }
        \nl \Return{$\log \tilde{p}(x) - \hat{U}$}\;
    \end{algorithm}
    
        
        
        
    }
    \end{minipage}\hfill%
    \begin{minipage}[t]{0.48\textwidth}
    {
        \removelatexerror
        \vspace{-10pt}
        \begin{algorithm}[H]
            \label{alg:alg7}
            \SetAlgoLined\DontPrintSemicolon
            \textbf{Algorithm 7:} RAVI EUBO estimator ($\texttt{EUBO}$)\;
            \KwIn{unnormalized model $\tilde{p}(x)$}
            \KwIn{exact sample $x \sim p(x)$}
            \KwIn{inference strategy $\mathcal{S}$ with arguments}
            \KwIn{arguments $\theta$}
            \KwOut{unbiased estimate of $\mathcal{U}$ (differentiable w.r.t. $\theta$)}
            \nl \uIf{$\mathcal{S}.q$ has a tractable marginal density}{
                \nl$\hat{L} \gets \log \mathcal{S}.q(x; \theta)$\;
            }
            \nl \ElseIf{$\mathcal{S}.q(x; \theta) = \int \mathcal{S}.q(r, x; \theta)\text{d}r$}{
                \nl $\hat{L} \gets \texttt{ELBO}(\mathcal{S}.q(\cdot, x; \theta), \mathcal{S}.\mathcal{M}, (x, \theta))$\;
            }
            \nl \Return{$\log \tilde{p}(x) - \hat{L}$}\;
        \end{algorithm}
    
        
    
        
    
            
    }
    \end{minipage}

    Note that in fact only every \textit{other} posterior approximation in the unrolled strategy requires a reparameterized version: Algorithm 7 never samples from its $\mathcal{S}.q$, only evaluates the densities.
 
    It would be interesting to develop variants of these algorithms that allow users to combine score-function and reparameterization estimation at different layers of nesting, or exploit other variance reduction tactics compositionally.

\end{document}